\documentclass[lettersize,journal]{IEEEtran}
\usepackage{amsmath,amsfonts,amssymb}
\usepackage{algorithmic}
\usepackage{bbding}
\usepackage{pifont}
\usepackage{wasysym}
\usepackage{algorithm}
\usepackage{array}

\usepackage{amsfonts}       
\usepackage{nicefrac}       
\usepackage{microtype}      
\usepackage{xcolor}

\usepackage{array} 
\usepackage{pifont}

\usepackage{bbding}

\usepackage[caption=false,font=normalsize,labelfont=sf,textfont=sf,singlelinecheck=false]{subfig}
\usepackage{caption}
\usepackage{textcomp}
\usepackage{multirow}
\usepackage{stfloats}
\usepackage{url}
\usepackage{verbatim}
\usepackage{graphicx}
\usepackage{cite}
\usepackage{colortbl}
\definecolor{mygray}{gray}{.9}
\definecolor{mygray2}{gray}{.8}
\definecolor{mypink}{rgb}{.99,.91,.95}
\definecolor{mycyan}{cmyk}{.3,0,0,0}
\usepackage{bbding}
\begin{document}

\title{DI-Net : Decomposed Implicit Garment Transfer Network for Digital Clothed 3D Human}

\author{Xiaojing Zhong, Yukun Su, Zhonghua Wu, Guosheng Lin$^*$, Qingyao Wu$^*$
\thanks{$^*$Corresponding author.}
\thanks{Xiaojing Zhong, Yukun Su and Qingyao Wu are with School of Software Engineering, South China University of Technology, Guangzhou, China. (email: vzxj12@gmail.com, suyukun666@gmail.com, qyw@scut.edu.cn)}
\thanks{Zhonghua Wu and Guosheng Lin are with School of Computer Science and Engineering, Nanyang Technological University, Singapore. (email: zhonghua001@e.ntu.edu.sg, gslin@ntu.edu.sg)}
}






\markboth{Journal of \LaTeX\ Class Files,~Vol.~14, No.~8, August~2021}%
{Shell \MakeLowercase{\textit{et al.}}: A Sample Article Using IEEEtran.cls for IEEE Journals}

\IEEEpubid{0000--0000/00\$00.00~\copyright~2021 IEEE}

\maketitle

\begin{abstract}
3D virtual try-on enjoys many potential applications and hence has attracted wide attention. However, it remains a challenging task that has not been adequately solved. Existing 2D virtual try-on methods cannot be directly extended to 3D since they lack the ability to perceive the depth of each pixel. Besides, 3D virtual try-on approaches are mostly built on the fixed topological structure and with heavy computation. To deal with these problems, we propose a Decomposed Implicit garment transfer network (DI-Net), which can effortlessly reconstruct a 3D human mesh with the newly try-on result and preserve the texture from an arbitrary perspective. Specifically, DI-Net consists of two modules: 1) A complementary warping module that warps the reference image to have the same pose as the source image through dense correspondence learning and sparse flow learning; 2) A geometry-aware decomposed transfer module that decomposes the garment transfer into image layout based transfer and texture based transfer, achieving surface and texture reconstruction by constructing pixel-aligned implicit functions. Experimental results show the effectiveness and superiority of our method in the 3D virtual try-on task, which can yield more high-quality results over other existing methods.


\end{abstract}

\begin{IEEEkeywords}
Human reconstruction, Pose transer, Human synthesis, Virtual try-on.
\end{IEEEkeywords}

\section{Introduction}
\IEEEPARstart{R}{ecently}, there has been a surge in researchers' interest in interpreting 3D contents. With the explosive growth of deep learning, a slew of related work has made significant progress \cite{bhatnagar2020loopreg,song20213d,bhatnagar2020combining,tiwari2021neural}. 3D virtual try-on is a novel and valuable task among them, which fits a specific clothing item onto a 3D human shape. Compared to image-based methods \cite{han2018viton,dong2019towards,yang2020towards,yu2019vtnfp,wang2018toward}, 3D virtual try-on is more close to life and more commercial, which can be widely applied in e-commerce and virtual games. However, the existing physics-based \cite{guan2012drape,hahn2014subspace} and scan-based \cite{lahner2018deepwrinkles,pons2017clothcap} methods face the challenges with high cost of data collection or computation. Despite learning-based methods \cite{bhatnagar2019multi,ma2020learning,mir2020learning} are convenient and amicable, most of them are subjected to fix topology structure since they construct the model on SMPL \cite{loper2015smpl}, which constrains the local expression to model folds and wrinkles. Despite Zhao $\emph{et~al.}$ \cite{zhao2021m3d} assigns the rgb values to front and back depth map of human to get colored point clouds, which ignores the consistent of whole texture. In general, several characteristics of human reconstruction constitute technical challenges for 3D virtual try-on from monocular images.
\par

The first challenge is that the relationship between garment images and reconstructed meshes is difficult to establish, as clothing often warps relative to perspective variations, and 2D images cannot provide unseen textures. A common solution is to build the UV mapping through projection and inpainting, as demonstrated in \cite{zhu2019detailed}, but this requires the mesh topology structure to remain fixed. Secondly, topology-free representations, such as regular voxel grids with high memory requirements and point clouds, are unsuitable for editing textures since they may lose structural information.

\par
Building upon recent progress in learning-based implicit representations for reconstruction with arbitrary topology and no resolution limitations \cite{mildenhall2021nerf,zheng2021pamir,chibane2020implicit,mescheder2019occupancy,park2019deepsdf}, this paper proposes a \textbf{D}ecomposed \textbf{I}mplicit garment transfer \textbf{Net}work (\textbf{DI-Net}) to solve the problem of reconstructing a 3D human model with a desired garment given monocular images that provide both human appearance and the target garment. The term "\textit{Decomposed}" carries two meanings in this context: first, we decompose the 3D virtual try-on process into transfers in the image and texture layouts, using the pixel-aligned feature embedding learned from the surface reconstruction to predict the per-vertex colors. Second, we further decompose the different attributes of human appearance based on human parsing maps and recombine them to perform garment transfer, which also indicates that our method can not only transfer the garment but any part of regions included in the parsing maps.

\par
\begin{figure}
\centering
\setlength{\abovecaptionskip}{0pt}
\includegraphics[width=0.48\textwidth,height=0.33\textheight]{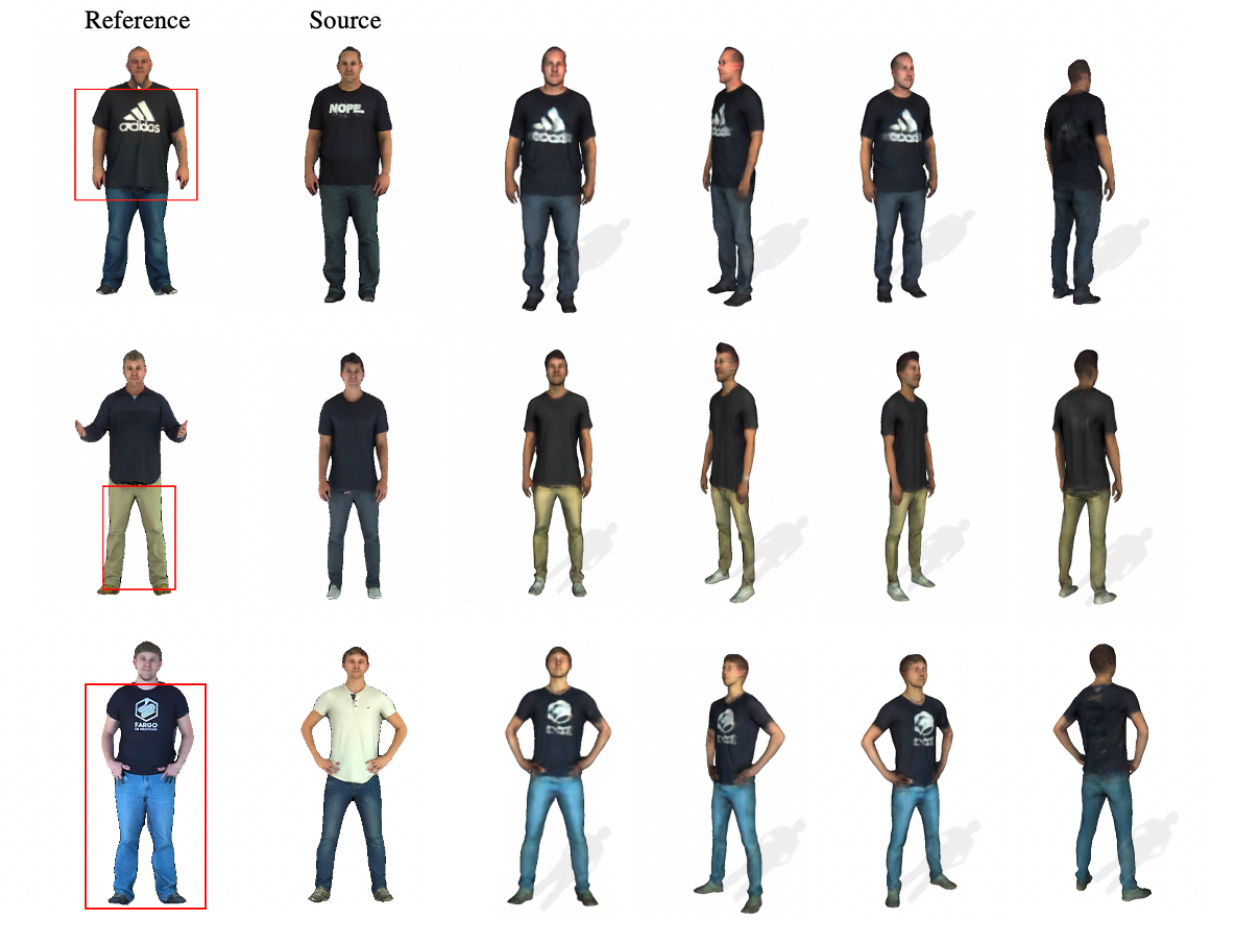}
\caption{\textbf{Illustrative examples of the proposed DI-Net.} Given the reference images that provide the target garments and the source images that provide the human appearance, our \textbf{DI-Net} can naturally reconstruct the 3D mesh with the desired garments from arbitrary perspectives. The highlighted areas indicate the garments that need to be transferred. Best viewed by zooming.}
\label{1}
\end{figure}

Specifically, we first introduce a complementary warping module to eliminate the spatial misalignment between the target image and the source image, which is responsible for warping the target image to have the same pose as the source image. Concretely, we propose to combine the advantages of both dense correspondence warping and sparse flow-based warping. The former provides accurate deformation positions but may fail to produce clear results since they are weighted by attention coefficients. The latter selects a local source patch for each output position to sample the pixels directly, despite the attention coefficient matrix being sparse. 
\IEEEpubidadjcol Then, we propose a geometry-aware decomposed transfer module, which consists of garment transfer on the image layout and texture layout. After transferring the garment in the image layout based on the warped image, we adopt pixel-aligned implicit functions to reconstruct the human shape with the transferred garment. In addition, for generating a consistent texture, we achieve texture layout garment transfer through fusing the feature maps from the source image and the warped image, estimating RGB values at each queried point. Fig. \ref{1} presents some results of our model and Tab. \ref{tab:1} presents an overview of the proper- ties of DI-Net and the most related approaches. Our contributions can be summarized as follows:
\begin{itemize}
\item{We address the 3D virtual try-on problem by decomposing the garment transfer onto the image layout and texture layout and predicting the position and color of each vertex based on the pixel-aligned implicit function. To the best of our knowledge, our method is the first attempt to reconstruct the human mesh with desired garments without any clothing or body templates.}

\item{To eliminate the spatial misalignment between the source image and the reference image for the following garment transfer, we propose a complementary warping module that combines the contributions of both dense correspondence learning and sparse flow learning to preserve details while achieving accurate warping.}

\item{Experiments demonstrate the superiority of our \textbf{DI-Net} in generating high-quality 3D virtual try-on results compared to these state-of-the-art methods.}

\end{itemize}

\section{Related works}
\subsection{2D and 3D Virtual Try-on.}
The goal of 2D virtual try-on methods is to place selected clothing items onto a model image \cite{du2022vton,hu2022spg,xu2021virtual}. In these methods, a clothing template is used, and non-rigid TPS transformation \cite{bookstein1991thin} is employed to warp the template so that it can fit the target pose. While some methods focus on modeling individual clothing items, our work is more closely related to those that aim to model all of a person's clothing simultaneously, allowing users to try on clothes from other person images without a clothing template. For instance, SwapNet \cite{raj2018swapnet} swaps clothing between a pair of images by disentangling the clothing from body shape and pose using segmentation masks. O-VITON \cite{neuberger2020image} separates appearance and shape generation and uses mutually exclusive segmentation masks to encode each component. Similarly, M2E \cite{wu2019m2e} and MV-TON \cite{zhong2021mv} transfer the desired clothes using region replacement with segmentation masks, but they also ensure that the model image has the same pose as the user image. ADGAN \cite{men2020controllable} embeds component attributes into the style code and reassembles them to render the person image, much like style transfer. CT-NET \cite{yang2021ct} steers the result away from absurd by merging correspondence learning between a pair of images and TPS warping, followed by dynamic fusion. DiO (Dressing in Order) \cite{cui2021dressing} estimates the global flow field instead of using TPS transformation to warp each component. Although 2D virtual try-on methods are expressive, they are restricted as they only play a role in the image. With the increasing demand for 3D virtual try-on, several works \cite{bhatnagar2019multi,mir2020learning,jiang2020bcnet} are devoted to explaining garments layered on 3D humans, building upon the parametric model SMPL \cite{loper2015smpl}. In these works, garments are defined as offsets of the vertices of the body mesh and rely on a template mesh. Textures are obtained through UV mapping according to the fixed topology. However, these methods tend to smooth results, and TailerNet \cite{patel2020tailornet} and Santesteban $\emph{et~al.}$ \cite{santesteban2021self} focus on retaining wrinkle detail of the mesh and predicting how the garments would fit in reality with a dynamic body. While using SMPL is easy to exhibit a human shape, it cannot handle complex topology, especially for people wearing dresses and skirts. M3D-VTON \cite{zhao2021m3d} predicts the depth map of a person image and matches the depth value with results of RGB-based virtual try-on, followed by triangulation to obtain a 3D clothed human. However, it only takes into account the view of the front and back, making the cohesion of the texture of the reconstructed human less smooth and unnatural.

\begin{table}
    \centering
    \scalebox{1.2}{
    \begin{tabular}{c|c|c|c|c}
        \hline
        \multirow{2}{*}{Method} & 
        \multicolumn{4}{c}{Property}\\
        \cline{2-5}& 2D & 3D &BT Free& CT Free \\
        \hline
        CP-VTON\cite{wang2018toward} & \ding{52} & &\ding{52} &  \\
        \hline
        ACGPN\cite{yang2020towards} &\ding{52} & &\ding{52} &  \\
        \hline
        ADGAN\cite{men2020controllable} &\ding{52} & &\ding{52} &  \ding{52}\\
        \hline
        MGN \cite{bhatnagar2019multi} & &\ding{52} &&  \\
        \hline
        NormalGAN\cite{wang2020normalgan} & & \ding{52}&\ding{52} &  \\
        \hline
        M3D-VTON \cite{zhao2021m3d}&\ding{52} & \ding{52} & \ding{52} \\
        \hline
        \rowcolor{mygray} DI-Net (\textit{Ours})&\ding{52} & \ding{52} & \ding{52} &\ding{52}\\
        \hline
    \end{tabular}
    }
    \vspace{2pt}
    \caption{Comparison of DI-Net to related work in terms of their properties. "2D": 2D virtual try-on. "3D": 3D human reconstruction. "BT Free": Body templates free. "CT Free": Clothing templates free.}\label{tab:1}
\end{table}

\begin{figure*}
\centering
\setlength{\abovecaptionskip}{0pt}
\includegraphics[width=1.0\textwidth,height=0.37\textheight]{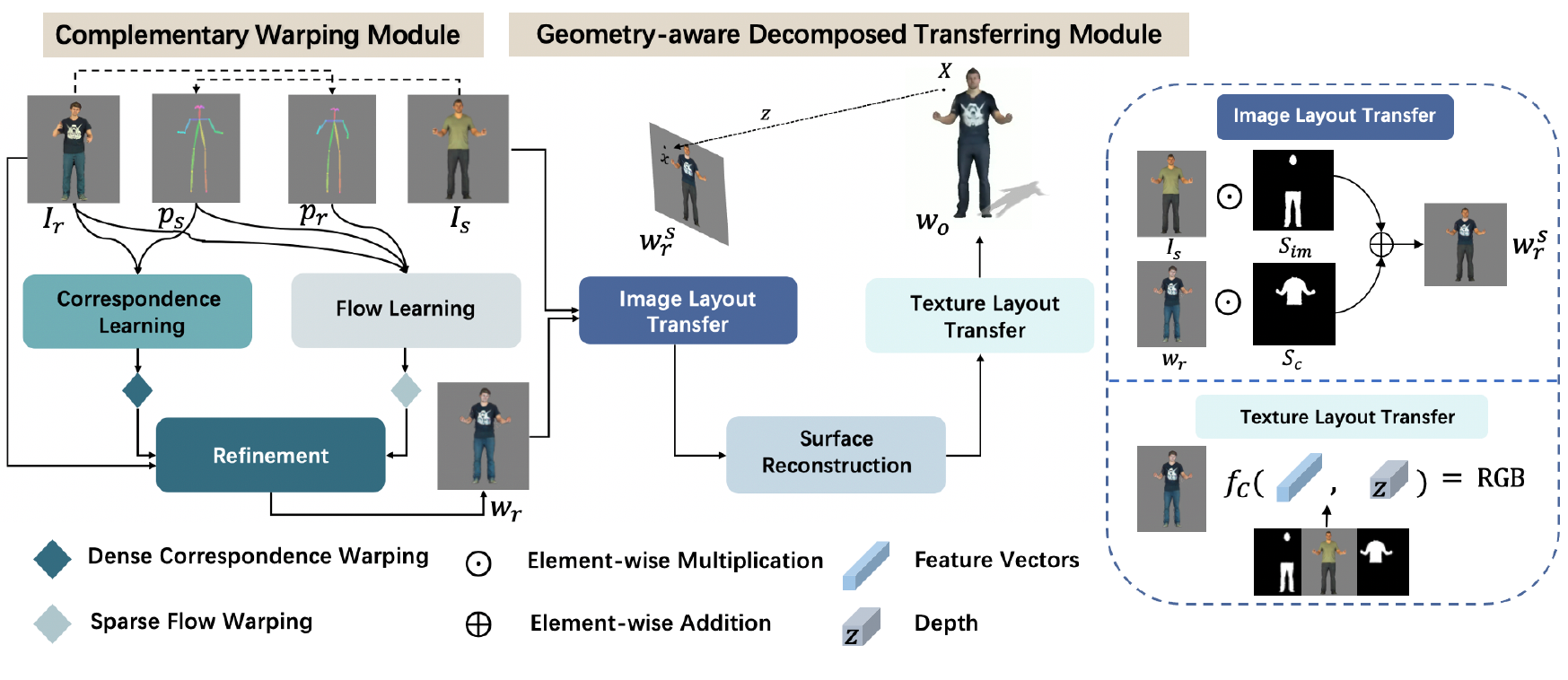}
\caption{\textbf{The overall architecture of our DI-Net.} (1) In Step I, given the reference image $I_r$, the source image $I_s$, the pose map $p_r$ of $I_r$, and the pose map $p_s$ of $I_s$, the Complementary Warping Module (\textbf{CWM}) combines the results through correspondence learning and flow learning to generate the warped image $w_r$, which has the same appearance as $I_r$ and the same pose as $p_r$. (2) In step II, the Geometry-aware Decomposed Transfer Module (\textbf{GDTM}) is introduced to decompose the garment transfer on 3D human into transfer on image layout and texture layout, where the former is based on the composition of $w_r$ and $I_s$ according to the corresponding masks, and the latter is established on the pixel-aligned implicit function that can be used to reconstruct the surface and texture of mesh simultaneously. $f_C(\cdot)$ is a implicit function that outputs the RGB value of each 3D vertex.  } 
\label{2}
\end{figure*} 

\section{Image-based Clothed Human Reconstruction.}
Recovering high-quality clothed 3D human mesh from monocular images is quite difficult due to challenges such as intrinsic uncertainties in lifting 2D observations to 3D space. In the field of garments, some works \cite{danvevrek2017deepgarment, guan2012drape, patel2020tailornet, gundogdu2019garnet, lahner2018deepwrinkles, santesteban2019learning} construct 3D garments as separate mesh layers, while others model the garments based on the prediction of displacement of body vertices \cite{bhatnagar2019multi, ma2020learning, zhu2019detailed, xiang2020monoclothcap, tiwari2020sizer}. The first generative model that produces 3D mesh with outfits from images directly is CAPE \cite{ma2020learning}, which uses a graph-based network to encode and decode the displacement. To make deformation more reasonable, HMD \cite{zhu2019detailed} applies a coarse-to-fine manner to project the mesh into 2D space with joint, anchor, and vertex-level handles, which allows the mesh to deform according to the movements of the handles. However, parametric models like SMPL \cite{loper2015smpl} and SMPL-X \cite{pavlakos2019expressive} have intrinsic limited power of representation to depict high-quality 3D clothed humans, despite their rendering efficiency and compatibility through predicting shape and pose parameters from images, as mentioned earlier. To support various topologies, Ma $\textit{et~al.}$ \cite{ma2021power,ma2021scale} model the garment using point clouds capable of exploiting explicit local representations. However, the performance depends on the gap consistency between patches. The implicit surface representation \cite{mescheder2019occupancy,park2019deepsdf} is also topologically flexible. By learning an implicit surface function, the goal is to decide whether a query 3D point is inside or outside of the shape at arbitrary resolutions. PIFu \cite{saito2019pifu} and its extension PIFuHD \cite{saito2020pifuhd} explore the alignment of 2D images with vertices of a 3D mesh, which is pivotal for reconstruction. The previous methods ignore the explicit alignment relationship between the two domains. PIFu adopts pixel-aligned features not only to reconstruct geometry but also to obtain texture. Nonetheless, it fails to change the texture, $\textit{e.g.}$ the ability to choose clothes derived from different images and demonstrate them on the textured mesh.

\section{Methodology}
In this paper, we aim to reconstruct a digital 3D model of a clothed human using monocular images and desired clothing. Given a reference image ($I_r$) depicting the desired clothing and a source image ($I_s$) providing human appearance and pose maps for both images, our goal is to reconstruct a 3D mesh ($w_o$) whose texture is composed of the desired clothes from $I_r$ and other parts from $I_s$. Notably, although the given images only offer a fixed viewpoint, the generated texture of the mesh is consistent and can be observed from any perspective. As shown in Fig. \ref{2}, our proposed method, DI-Net, consists of the Complementary Warping Module (CWM) and the Geometry-Aware Decomposed Transfer Module (GDTM). Intuitively, since prior clothing templates are not provided, we separate the component attributes of the human and then recombine the desired attributes while ensuring spatial alignment.

\begin{figure}
\centering
\setlength{\abovecaptionskip}{0pt}
\includegraphics[width=0.48\textwidth,height=0.23\textheight]{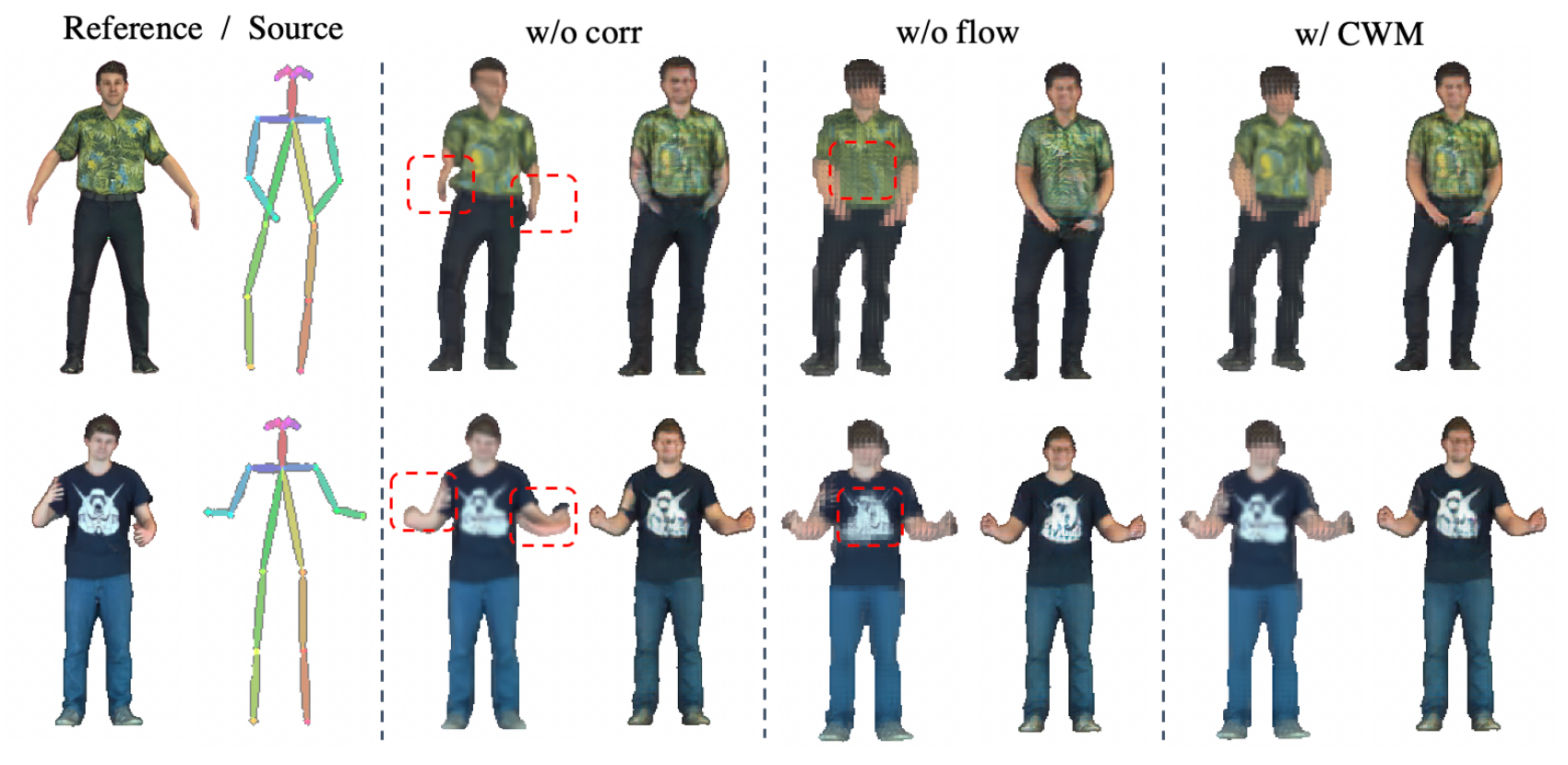}
\caption{"\textit{w/o corr}" denotes warping the garment without correspondence learning and "\textit{w/o flow}" denotes without flow learning. From the second column, the left and right represent the generated results before and after the refinement module respectively. The highlighted areas indicate the blurred or unnatural texture caused by dense correspondence warping and the cracked or distorted torsos caused by flow warping. Note that we only need the generated garment of this module, so the distortion of the face is acceptable.}
\label{3}
\end{figure}

\subsection{Complementary Warping Module}

Warping the clothing extracted from the reference image to fit the source image tends to be more difficult than warping the clothing templates since they are often frontal and erratic. Tps warping \cite{han2018viton,wang2018toward,yang2020towards} with a low degree of freedom fails to perform well when facing complex scenarios. We consider that the model should support flexible spatial manipulations of the source image to deal with different parts. In particular, clothing is required to preserve the fine details during the warping, while body appearance has a dull texture but may fail to warp to the correct position, as shown in Fig. \ref{3}. Therefore, we propose to warp body appearance through learning dense correspondence between the source image and the target pose since long-range correlation contributes to shaping large geometric changes. Additionally, to preserve the details of clothing, we warp it by estimating the flow field that has an explicit one-to-one mapping between each output position and its respective input location. Finally, we treat the combination of the warped results as the input to a refinement module, which will progressively synthesize the output $w_r$.

\subsubsection{Correspondence Learning}
We adopt $I_r$ and $p_s$ as inputs, where $p_s$ is extracted by an off-the-shelf pose estimator \cite{cao2017realtime} and further converted into distance keypoint fields as described in \cite{zhang2020cross}. Specifically, we first provide two independent feature extractors $\mathcal{F}_r$ and $\mathcal{F}_s$ to embed the features extracted from $I_r$ and $p_s$ into the same domain. We denote the outputs as $f_r\in \mathbb{R}^{H \times W \times C}$ and $f_s\in \mathbb{R}^{H \times W \times C}$, respectively. This can be formulated as:

\begin{equation}
\begin{aligned}
f_r = \mathcal{F}_r(I_r) \\
f_s = \mathcal{F}_s(p_s),
\end{aligned}
\end{equation} 

Following this, we propose to leverage cosine similarity to match the aggregated features $f_r^{'}$ and $f_s^{'}$, since it has been proven useful in calculating pixel-wise semantic relevance \cite{zhang2019deep, luo2018cosine}. The correspondence matrix $M\in \mathbb{R}^{HW \times HW}$ can then be computed as:


\begin{equation}
M(i,j)= \frac{(f_r^{'}(i)^T-u_r)(f_s^{'}(j)-u_s)}{\Vert(f_r^{'}(i)-u_r)\Vert \Vert(f_s^{'}(j)-u_s)\Vert},
\end{equation}
where $u_r$ and $u_s$ represent the mean vectors. $i$ and $j$ represent $f_r^{'}(i)$ at position $i$ and $f_s^{'}(j)$ at position $j$ respectively, used to calculate channel-wise centralized features.
\par
Accordingly, the dense correspondence warping $W^d\{\cdot\}$ can be obtained by:
\begin{equation}
W^d_{I_r}(u)= \sum\limits_{v}\mathop{softmax}\limits_{v}(\alpha M(u,v) \cdot I_r(v)),
\end{equation} 
where $\alpha$ is a hyper-parameter to control the sharpness of the results and is set as 100 empirically. We utilize a human parsing algorithm \cite{wei2017learning} to obtain the parsing map $S\in \mathbb{R}^{H \times W \times 20}$ of $I_r$, where 20 denotes the number of labels. We also warp $S$ to extract the body appearance of $W^d_{I_r}$, except for the clothing, and mark it as $W^d_{I_h}$.

\subsubsection{Flow Learning}
We employ a flow estimator to predict the motions between $p_r$ and $p_s$, which yields the global flow filed $f_w$ at different scales. It can be formulated as follows:


\begin{equation}
f_w = \mathcal{F}(I_s,p_r,p_s),
\end{equation}
where $\mathcal{F}$ is a fully convolutional network and $f_w$ indicates a sparse attention coefficient matrix that specifies which pixels could be sampled from the local patch. Therefore, the sparse flow warping $W^f{\cdot}$ can be obtained according to the coordinate offsets at the maximum scale as follows:

\begin{equation}
W^f_{I_r}(u)=f_w(I_r(v)), 
\end{equation}

In this part, we separate the clothing of $W^f_{I_r}$ to acquire $W^f_{I_c}$ using the warped parsing map. Additionally, as the labels of the flow fields are often not available, we constrain $f_w$ in an unsupervised manner.

\par
\textbf{Sampling correctness loss.} Following \cite{ren2019structureflow}, we denote the feature maps extracted from the warped image $W^f_{I_r}$ and the ground-truth image by a VGG network as $v_r$ and $v_t$ respectively. We attempt to make the generated flow fields to sample positions with similar semantics through measuring the similarity between $v_r$ and $v_t$.

\begin{equation}
 \mathcal{L}_{flow}=
\sum_{l\in \eth }exp(-\frac{cos(v_r(k),v_t(k))}{\mu}),
\end{equation}
where the coordinate set $\eth$ contains all positions and $l$ represents the location. $\mu$ is a normalization term used to avoid the bias brought by occlusion."

\par
\textbf{Regularization term.} To capture the highly correlated deformations of image neighborhoods (like arms and clothes), we incorporate a regularization term into the flow fields. This term penalizes non-affine transformations in local regions, thereby enhancing the network's ability to capture the close relationship between deformations. 
\begin{equation}
 \mathcal{L}_{regular}=
\sum_{l\in \eth }|| T_l - \theta S_l ||_2^2,
\end{equation}
where $T_l$ denotes n $\times$ n patch of the target features with the location $l$ and $S_l$ denotes the source features. The estimated affine transformation term $\theta$ can be obtained by using the least-squares estimation as $\theta = (S_l^HS_l)^{-1}S_l^HT_l$.

\subsubsection{Refinement Module}

Afterward, we are able to get the combined warped image $w_r^{\prime}$ as follows:
\begin{equation}
w_r^{\prime} = W^d_{I_h} + W^f_{I_c}.
\end{equation}

Guided by the garment of $w_r^{\prime}$, we refine $w_r^{\prime}$ to obtain the final warped result $w_r$ by scaling and shifting the feature maps gradually with the learned modulation parameters in the spatially-adaptive denormalization (SPADE) \cite{park2019semantic} architecture. To clarify, the proposed method combines a warped image from the complementary warping module and the keypoint distance of the source image, concatenated along the channel-wise dimension, as intermediates. The goal of the refinement module is to refine the combined warped image to produce fine-grained results based on the intermediates. The intermediates are first reduced to an 8x8 size, and the refinement module consists of seven resblocks, which are normalization layers with an accompanying upsample layer. The number of channels is adjusted using convolutional layers at the start and end of the resblocks. Each resblock outputs the scale and bias parameters $\alpha_i$ and $\beta_i$ to modulate the normalization layer, with the intermediates resized to have the same shape as the feature maps sent to the layer. Additionally, the method uses a discriminator from pix2pixHD \cite{wang2018high}.


\begin{figure}
\centering
\setlength{\abovecaptionskip}{0pt}
\includegraphics[width=0.48\textwidth,height=0.23\textheight]{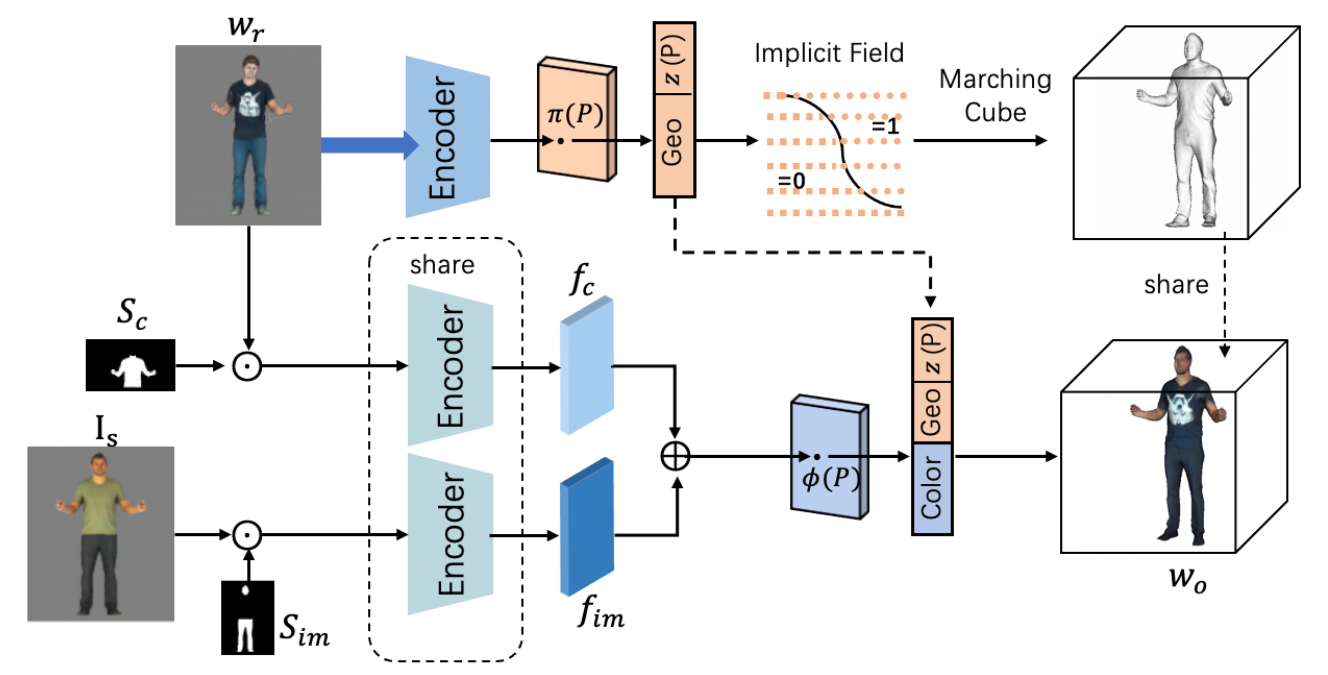}
\caption{\textbf{The framework of the texture layout transfer of the geometry-aware decomposed transfer module.} $\oplus$ means element-wise plus and $\odot$ denotes element-wise multiply.}
\label{4}
\end{figure}

\section{Geometry-aware Decomposed Transfer Module}
To better perceive the geometrical structure of each pixel, we propose to decompose the 3D garment transfer into two parts: image layout and texture layout. By utilizing the result of the image layout transfer as the input for shape reconstruction, we can provide the reconstructed human with a new topological representation while wearing the desired garments. In addition, we view the surface texture as a vector function that is defined in the space close to the surface. This approach enables the creation of textures for digital humans with freely-specified topology and self-occlusion, resulting in more realistic and visually appealing digital models.
\subsection{Image layout transfer}

Here, $w_r$ represents the warped image that has the same pose as $I_s$ while preserving the appearance of $I_r$, which is generated by the complementary warping module. Note that the appearance may change due to the occlusion or exposure caused by the garment transfer. To address this, we transfer the warped garment and arms simultaneously. As shown in Figure \ref{2} (top right), we can obtain the image layout transfer result $w_r^s$ as follows:

\begin{equation}
w_r^s = (I_s \odot S_{im})\oplus(w_r \odot S_c),
\end{equation}
where $S_{c}$ means the arms and clothes mask of the warped parsing map. $S_{im}$ denotes the other part except them.

\subsection{Texture layout transfer}

Fig. \ref{4} shows the framework of how texture layout transfer works. First, we use the pixel-aligned implicit function \cite{saito2019pifu} to reconstruct the human shape after getting $w_r$ from the image layout transfer. Specifically, a 3D surface can be defined as a level set of function $f$, e.g. $f(X)$ =0.5, where X is a 3D point and $f:\mathbb{R}^3\rightarrow[0,1]$ is represented by a deep neural network. To infer the 3D textured surface from a single image, the pixel-aligned feature is utilized as the condition variable. Thus, $f$ can be extended as follows:
\begin{equation}
f(F(x),z(X)) = s:s\in \mathbb{R},
\end{equation}
where $x=\phi(X)$ gives the 2D images projection point of X. F(x)=g(I(x)) is the local image feature at x extracted by a fully covolutional image encoder g, and z(X) is the depth value in weak-perspective camera coordinate. $I\{\cdot\}$ is a sampling function used to sample the value of the feature map at pixel $\phi(x)$ using bilinear interpolation. PIFu takes a given pixel and uses the local image feature of the pixel to cast a ray along the z-axis and estimate the values of occupancy probability along that ray. During inference, the iso-surface of the probability field is recovered using the Marching Cube \cite{lorensen1987marching}.

Instead of a scalar field, the output of $f(\cdot,\cdot)$ is specified as an RGB vector field, which can be used to predict the color of each vertex when given a single image. To achieve this, we design decomposed component encoders that embed the attributes into the geometry-aware latent space. This space is then sampled by $I{\cdot}$ to provide consistent RGB values. Note that the semantic labels used by the decomposed component encoders are flexible, allowing our DI-Net to be applied to both top and bottom clothing transfers. The texture layout transfer operation is defined as follows:

\begin{equation}
\begin{aligned}
f_C(F_c(x),z(X),F_g(x)) & = RGB:RGB\in \mathbb{R}\\
    F_c(x) & = (f_c \oplus f_{im})(I(x)),
\end{aligned}
\end{equation} 
where $f_c$ and $f_{im}$ are the high-level features extracted from $w_r$ and $S_{im}$ according to $S_c$ and $S_{im}$, respectively. Besides, $F_g(x)$ is the feature embedding learned from the shape reconstruction, which make $f_C$ capable of inferring the texture of an unseen surface. 

\subsection{Training Losses}

The training process involves achieving two goals: pose transfer with $\mathcal{L}_{perc}$, $\mathcal{L}_{ctx}$, $\mathcal{L}_{adv}$, and $\mathcal{L}_{cycle}$ in the complementary warping module; and shape reconstruction and texture prediction with $\mathcal{L}_{regS}$ and $\mathcal{L}_{regC}$ in the geometry-aware decomposed transfer module.

\par
\textbf{Perceptual loss.} We employ the perceptual loss \cite{johnson2016perceptual} to constrain the high-level semantic similarity between $w_r$ and the ground truth $w_t$. We utilize VGG-19 pretrained model\cite{simonyan2014very} to extract multi-level features $\phi_l$ and calculate $\mathcal{L}_{perc}$ as follows:
\begin{equation}
\mathcal{L}_{perc} = ||\phi_l(w_r)-\phi_l(w_t)||_2.
\end{equation}

\textbf{Contextual loss.} To penalize the semantically mismatch between $w_r$ and $I_r$, we adopt the contextual loss proposed in \cite{mechrez2018contextual} to preserve more details during the generation. $\mathcal{L}_{ctx}$ can be obtained as follows:
\begin{equation}
\begin{aligned}
\mathcal{L}_{ctx} =\sum\limits_{l}\omega_l[-log(\frac{1}{nl}\sum\limits_{i}\mathop{max}\limits_{j}A^l(\phi^l_i(w_r),\phi^l_j(w_t)))], 
\end{aligned}
\end{equation}
where $A^l$ denotes the pairwise affinities between features.

\textbf{Adversarial loss.} To make $w_r$ close to the real distributions of $w_t$, we leverage a discriminator \cite{mirza2014conditional} to constrain the output space of the generator. $\mathcal{L}_{adv}$ can be formulated as follows:

\begin{equation}
\begin{aligned}
\mathcal{L}_{adv}(G,D)& = \mathbb{E}_{I_r,p_s}[log(1-D(G(p_s,I_r)|p_s,I_r)]\\
&+ \mathbb{E}_{I_r,p_s}[logD(w_t|p_s,I_r)],
\end{aligned}
\end{equation}

\textbf{Cycle loss.} There lacks a powerful supervision to learn the correspondence matrix. To solve that, we construct a cyclic warping as \cite{zhang2020cross} to ensure the learned correspondence is cycle-consistent. We warp $w_r$ again with the correspondence matrix to compare with $I_r$. $\mathcal{L}_{cycle}$ can be computed as follows:
\begin{equation}
\mathcal{L}_{cycle}=||\hat{w}_r-I_r||_1,
\end{equation}
where $\hat{w}_r$ is the result derived from warping $w_r$.
\par
\textbf{Regression loss.} 
For surface and texture reconstruction, we directly adopt the regression loss to learn the mapping.
\begin{equation}
\begin{aligned}
& \mathcal{L}_{regS}=\frac{1}{n}\sum\limits_{i=1}^n||s,f^*(X_i)||_2 \\
& \mathcal{L}_{regC}=\frac{1}{n}\sum\limits_{i=1}^n||r,f^*_C(X_i)||_2,
\end{aligned}
\end{equation}
where $s$ is the output of $f(\cdot,\cdot)$ and r is the output of $f_C(\cdot,\cdot)$. $f^*(X_i)$ and $f^*_C(X_i)$ are their ground truth respectively.
\par
Finally, the overall loss can be formulated as follows:
\begin{equation}
\begin{aligned}
\mathcal{L}_{full} &=  \lambda_1\mathcal{L}_{flow}+\lambda_2\mathcal{L}_{regular}+\lambda_3\mathcal{L}_{perc}+\lambda_4\mathcal{L}_{ctx}+\\
&\lambda_5\mathcal{L}_{adv}+\lambda_6\mathcal{L}_{cycle}+\mathcal{L}_{regS}+\mathcal{L}_{regC}
\end{aligned}
\label{eq16}
\end{equation}
where $\lambda_1$, $\lambda_2$, $\lambda_3$, $\lambda_4$, $\lambda_5$ and $\lambda_6$ are the trade-off parameters. $\mathcal{L}_{regS}$ and $\mathcal{L}_{regC}$ are trained separately.

\par

\begin{figure}
\centering
\setlength{\abovecaptionskip}{0pt}
\includegraphics[width=0.48\textwidth,height=0.3\textheight]{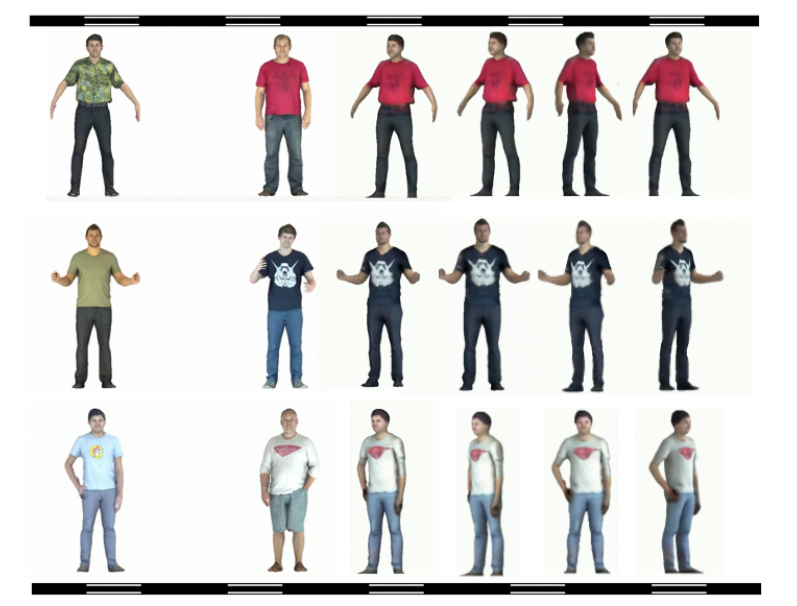}
\caption{\textbf{More rendered results on novel views.}}
\label{5}
\end{figure}

\section{Experiments}

\begin{figure*}
\centering
\setlength{\abovecaptionskip}{0pt}
\includegraphics[width=1.0\textwidth,height=0.47\textheight]{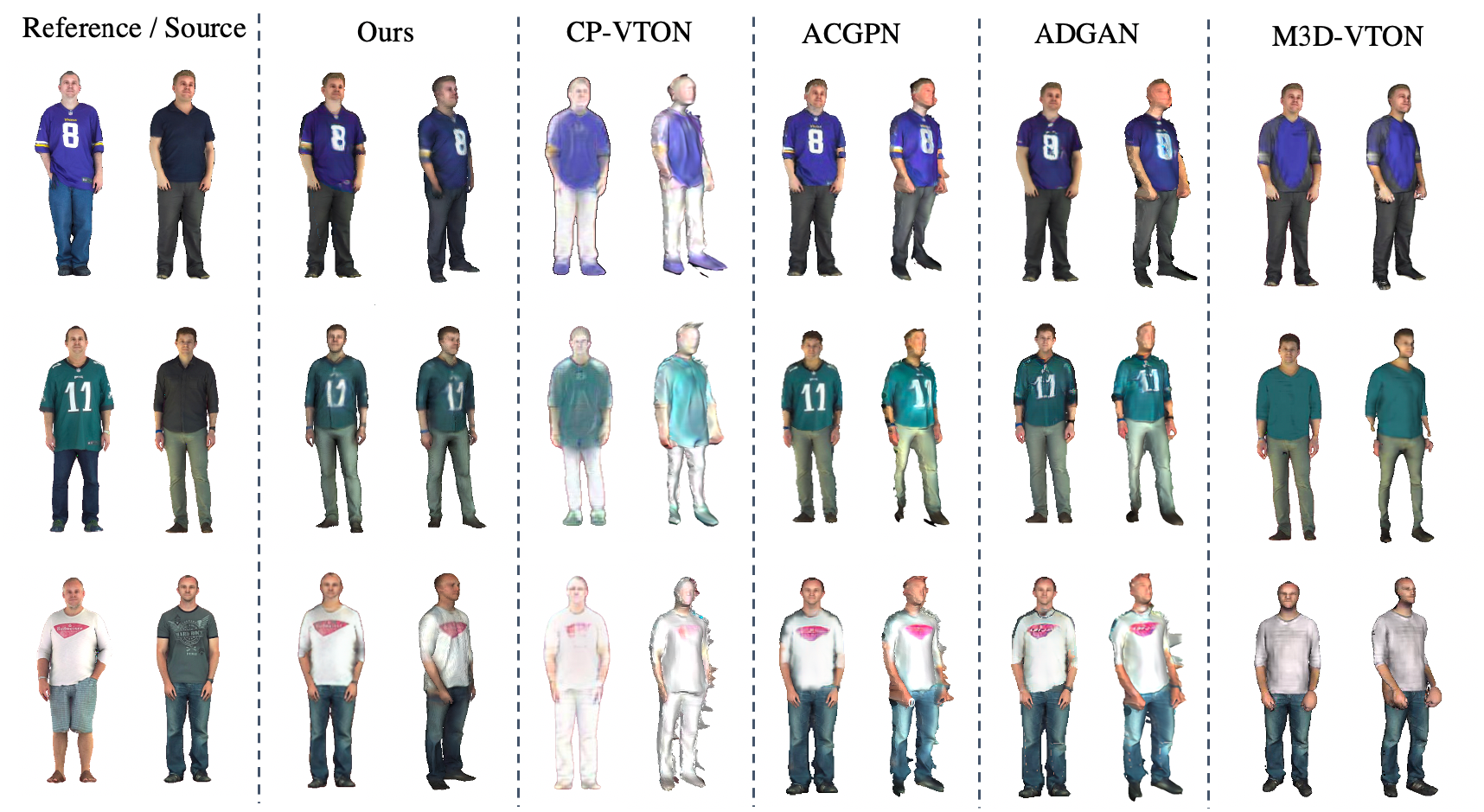}
\caption{\textbf{Qualitative comparisons of 2D and 3D try-on results.} We split the images into five columns. The first columns represent the reference and the source image. In each other column, the left and right shows the 2D and 3D try-on results, respectively. Among them, CP-VTON \cite{wang2018toward}, ACGPN\cite{yang2020towards} and ADGAN\cite{men2020controllable} are 2D-based methods thus they perform 3D virtual try-on using NormalGAN \cite{wang2020normalgan}.} 
\label{6}
\end{figure*}

\subsection{Dataset} 
In order to learn 3D reconstruction like \cite{saito2019pifu,saito2020pifuhd}, our method needs access to textured mesh. There are several restrictions on the datasets we can utilize to train our model: 1) High-quality textured mesh with well-polished by artists is private and commercial like \cite{renderpeople}. 2) datasets with complex poses \cite{zheng2019deephuman,yu2021function4d} are not suitable for our model since we need to extract the garments from the reference image. Therefore, we opt for the MGN digital wardrobe \cite{bhatnagar2019multi} as our training dataset. MGN merely releases 96 scans of persons dressed differently rather than 356 noticed in their paper. We randomly select 83 of them for training and the rest for testing. Similar to \cite{saito2019pifu,saito2020pifuhd}, we apply a Lambertian diffuse shader and spherical harmonic lighting \cite{varol2017learning} to render images with a weak perspective camera and rotate each scanned people at every four angles. Images are displayed at a resolution of 512 x 512. After completion, we are able to collect 7,470 rendered images for training. Each scanned person corresponds to 90 perspectives and we select 2 different of them to construct a train pair. We arbitrarily select 80 pairs for each scanned person. An image corresponds to a mesh when train the geometry-aware decomposed transferring module. Further results regarding novel views in the MGN dataset can be seen in Fig. \ref{5}. In addition, we also utilize DeepFashion dataset \cite{liu2016deepfashion} to validate the effectiveness of our complementary warping module, which contains 52712 high-quality model images with clean backgrounds. 


\subsection{Implementation Details}

The CWM and GDTM are trained separately. To train the CWM, we first train the flow learning for 20 epochs to estimate reasonable flows and then jointly train it with the correspondence learning and refinement module in an end-to-end manner. Training the GDTM is divided into two steps: geometry reconstruction and texture reconstruction. We adopt the point sampling scheme proposed in \cite{saito2019pifu}, which combines uniform sampling and adaptive sampling based on the surface geometry and utilizes the Embree algorithm for occupancy querying. We add noise to the surface points along the normal so that the color can be defined not only on the exact surface but also in the 3D space around it. All the code is implemented using the deep learning toolkit PyTorch, and a single NVIDIA 2080ti GPU is used in our experiments. To balance the scales of losses in Equation \ref{eq16}, we empirically set $\lambda_{1,2,3,4,5,6}$ to 1.0, 1.0, 0.001, 1.0, 10.0, and 100.0, respectively.

\begin{figure}
\centering
\setlength{\abovecaptionskip}{0pt}
\includegraphics[width=0.48\textwidth,height=0.12\textheight]{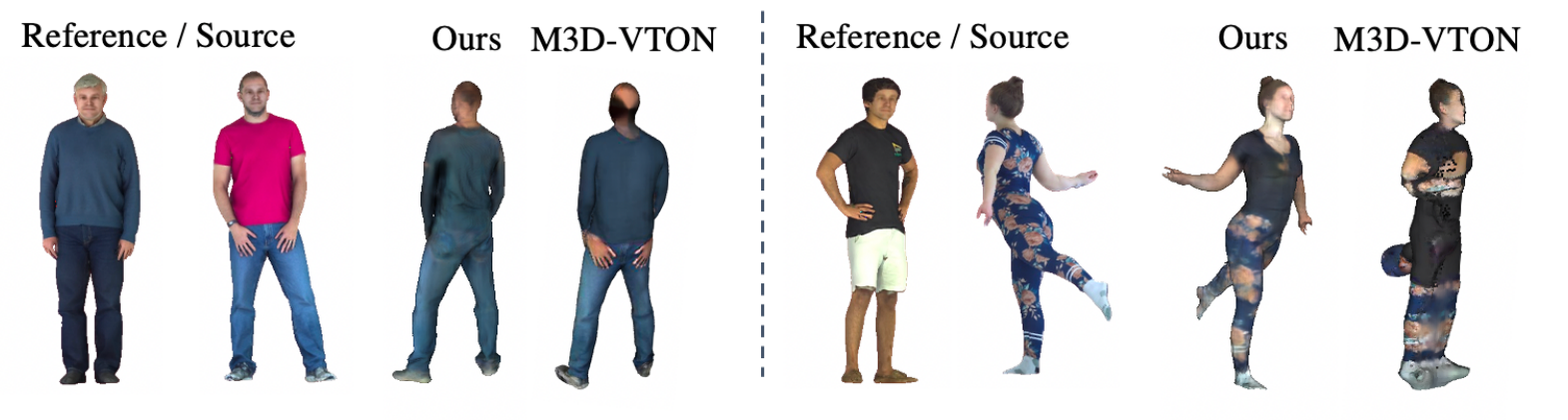}
\caption{\textbf{Comparison of our method and M3D-VTON in large spatial rotation.}}
\label{7}
\end{figure}

\subsection{Qualitative Results}
\textbf{MGN.} We perform a visual comparison of our proposed method with CP-VTON \cite{wang2018toward}, ACGPN \cite{yang2020towards}, ADGAN \cite{men2020controllable} and M3D-VTON \cite{zhao2021m3d}. As illustrated in Tab \ref{tab:1}, CP-VTON and ACGPN are 2D-based methods that require clothing templates, while ADGAN is devoid of the need for a clothing template; M3D-VTON is able to realize 3D virtual try-on with clothing templates. For a multi-view comparison, we use NormalGAN \cite{wang2020normalgan} to produce the textured mesh based on the results of 2D try-on approaches. All of these approaches are retrained using the same training set as ours on the MGN \cite{bhatnagar2019multi} dataset to ensure that our trials are comparable. Fig. \ref{6} provides a qualitative comparison. The appearance of the source image is difficult to keep using CP-VTON, and it can only transfer the color of the desired garment while losing most of the texture. ACGPN enhances TPS warping with STN \cite{jaderberg2015spatial}, making it superior to CP-VITON. During the warping process, ACGPN maintains the pattern and texture, but the generated limbs are not as natural looking as the ones we get, like the arms in the second row. At the same time, the edges of some garments are in shadow, as shown in the third row. ADGAN is designed for template-free garment transfer; it does this by encoding decomposed human attributes into a latent space. From the results, the garment's texture is maintained. However, it tends to lose tiny details such as the sleeves, as shown in the first and third rows. In addition, the pattern of the warped garment is somewhat distorted in scale. M3D-VTON is the first attempt to reconstruct a 3D try-on textured mesh using monocular images as inputs. During the transfer, it can obtain a smooth texture but lose the pattern. In addition, it is feasible to see negative cases such as broken arms. As seen in Fig. \ref{7}, the back texture generated by M3D-VTON has hands that should not be there due to the fact that M3D-VTON obtains the back texture via mirror-mapping. Contrastingly, our approach not only entirely restores the identity feature from the source image but also transfers desired garments onto the target person in a way that looks seamless and consistent.
\par
\begin{figure}
\centering
\setlength{\abovecaptionskip}{0pt}
\includegraphics[width=0.4\textwidth,height=0.42\textheight]{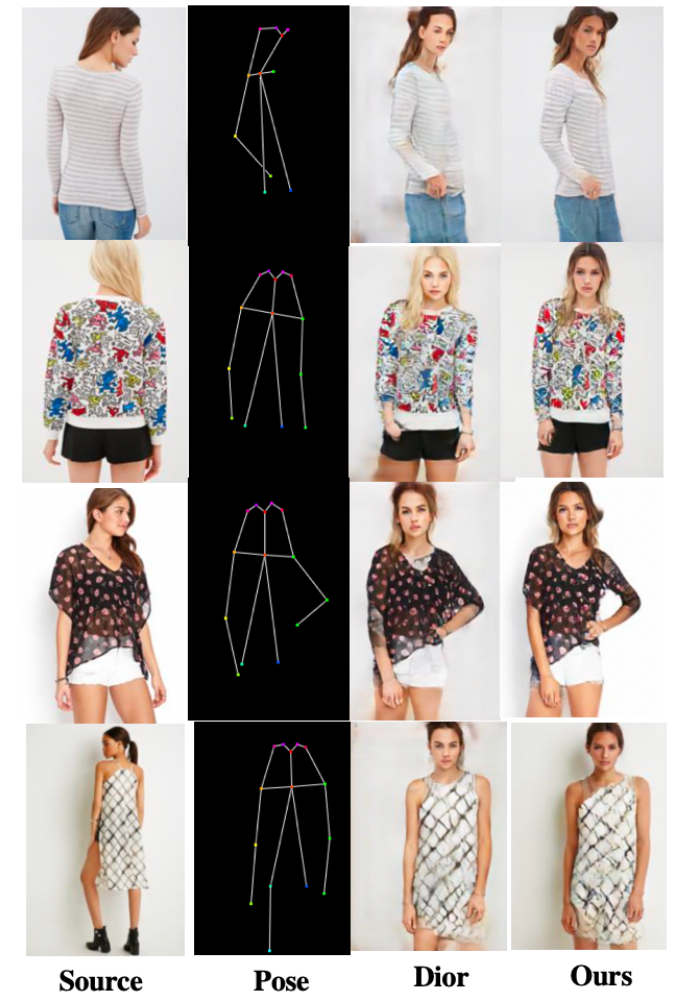}
\caption{\textbf{Comparison of our complementary warping module and Dior\cite{cui2021dressing} regarding pose transfer in DeepFashion dataset.} It can be seen that CWM can generate clear texture and body}
\label{8}
\end{figure}

\textbf{DeepFashion.} We compared our pose transfer method for CWM in 2D implementation with the DIOR method on the Deepfashion dataset, where complex poses including side and back transformations were involved. As shown in Fig. \ref{8}, due to the combination of correspondence learning in CWM, it can avoid distortion and blurring of limbs during the transformation process. The effect of flow learning makes the texture of the generated image clearer, which can provide clear texture and appearance for the subsequent 3D steps.

\begin{table}
    \centering
    \scalebox{1.2}{
    \begin{tabular}{c c c c }
        \hline
        Method & SSIM  $\uparrow$ & FID  $\downarrow$ & LPIPS $\downarrow$ 
        \\
        \hline
        CP-VTON\cite{wang2018toward} & 0.4197 & 267.3679 &  0.4356\\
        ACGPN\cite{yang2020towards}& 0.9162 & 85.5888 &  0.0799\\
        ADGAN\cite{men2020controllable}& 0.9080 & 67.9037 &  0.0496\\
        M3D-VTON \cite{zhao2021m3d}& 0.9094 & 92.9363 &  0.0732\\
        \hline
        \rowcolor{mygray} DI-Net (w/flow) & 0.9700 & 45.1268 &  0.0126\\
        \rowcolor{mygray} DI-Net (w/corr) & 0.9698 & 49.9880 &  0.0130\\
        \hline
      \rowcolor{mygray2} DI-Net (w/o $\mathcal{L}_{perc}$) & 0.9694 & 47.5125 & 0.0128\\
        \rowcolor{mygray2} DI-Net(w/o $\mathcal{L}_{ctx}$)& 0.9703 & 43.6480 & 0.0120 \\
         \rowcolor{mygray2} DI-Net(w/o $\mathcal{L}_{cycle}$) & 0.9708 & 43.1388 &  0.0118\\
        \rowcolor{mygray2} DI-Net (full) & \textbf{0.9714} & \textbf{42.9196} & \textbf{0.0109}\\
        \hline
    \end{tabular}}
    \vspace{2pt}
    \caption{Quantitative comparisons of our method with other methods in terms of SSIM \cite{wang2004image}, FID \cite{heusel2017gans} and LPIPS \cite{zhang2018unreasonable}. Besides, DI-Net (w/o $\mathcal{L}_{perc}$) denotes without perceptual loss; DI-Net (w/o $\mathcal{L}_{ctx}$) denotes without contextual loss; DI-Net (w/o $\mathcal{L}_{cycle}$) denotes without cycle loss. DI-Net (w/flow) denotes without correspondence learning. DI-Net (w/corr) denotes without flow learning. DI-Net (full) denotes using the full model. They are DI-Net variants for ablation study.}
    
    \label{tab:2}
\end{table}

We quantitatively evaluate the effectiveness of our method for image-layout transfer using three commonly used metrics: SSIM, FID, and LPIPS. SSIM \cite{wang2004image} compares the luminance, contrast, and structure components of two images to calculate their structural similarity index \cite{zhou2021semantic, pang2021image, lin2023smnet}. To compute SSIM, we first calculate the mean and variance of each component for both images and then compute the product of three terms, each representing the similarity between the corresponding components of the two images. Fr'{e}chet Inception Distance (FID) \cite{heusel2017gans} is a metric commonly used to evaluate the realism of generated images in the field of generative adversarial networks (GANs). FID calculates the Wasserstein-2 distance between the feature representations of the generated and ground truth images obtained from a pre-trained deep convolutional neural network (CNN), such as Inception V3. LPIPS (Learned Perceptual Image Patch Similarity) \cite{zhang2018unreasonable} assesses the visual quality of generated images based on human perception, using a deep neural network to learn a feature space that captures perceptual similarity between images. In Tab. \ref{2}, we present the scores obtained by our method and compare them with those obtained by other methods. Our method achieves the highest scores in both SSIM and LPIPS, which indicates that it performs well in preserving the details and structural characteristics of real images. Our method also achieves the lowest scores in FID, suggesting that the generated images are visually very similar to the real images.


\subsection{Ablation Study}

\textbf{Effectiveness of the Complementary Warping Module (CWM).} Visually, sparse flow warping provides sharp sampling, which aids in preserving textures during transfer, while dense correspondence warping ensures that the body is transferred to the correct position (see Fig. \ref{3}). The combination of these two modules benefits the refinement module in generating complete body and photorealistic appearance results compared to individual modules. Based on the metrics presented in Tab. \ref{2}, it is evident that the full model generates better images compared to the model without flow warping or dense correspondence warping.
\par 

\begin{figure}
\centering
\setlength{\abovecaptionskip}{0pt}
\includegraphics[width=0.48\textwidth,height=0.23\textheight]{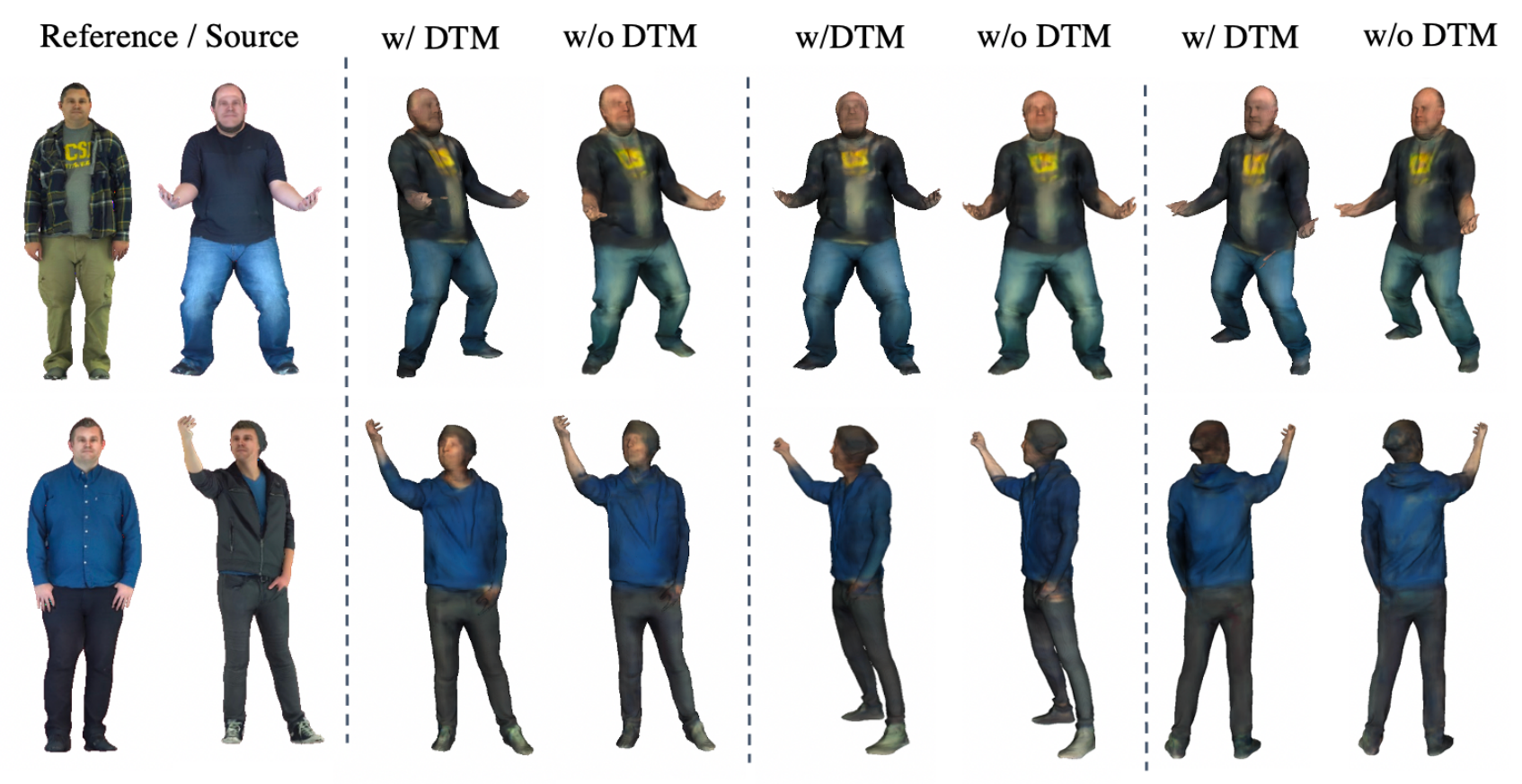}
\caption{\textbf{Visual comparisons of multi-views to verify the effectiveness of the Decomposed Transfer Module (DTM).}}
\label{9}
\end{figure}

\textbf{Effectiveness of the Geometry-aware Decomposed Transfer Module (GDTM).} GDTM captures geometric features attached to each pixel, resulting in cohesive and consistent texture. To prove the efficacy of GDTM, we provide results of multi-views with and without GDTM in Fig. \ref{9}. The texture generated without GDTM in the first row shows both long and short sleeves in different views, as it cannot perceive semantic information spatially with geometry awareness. Additionally, the connection between clothes and pants is not as natural as the results generated by GDTM. Artifacts are also visible at the junction of the neck and clothing in the second row. 

\begin{figure}[htbp]
\centering 
\setlength{\abovecaptionskip}{0pt}
\includegraphics[width=0.3\textwidth,height=0.2\textheight]{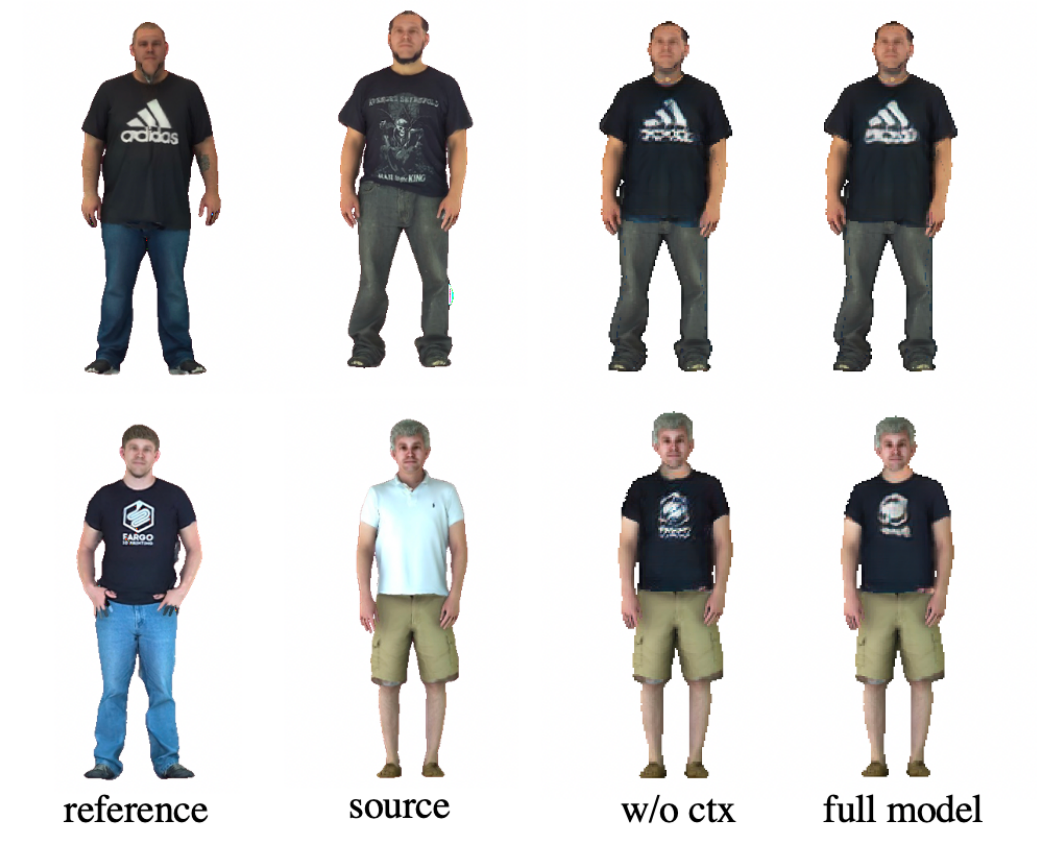}  
\caption{\textbf{Effects of the contextual loss.} \textit{w/o ctx} indicates that training our method without using contextual loss.}
\label{10}
\end{figure}

\begin{figure}[htbp]
\centering 
\setlength{\abovecaptionskip}{0pt}
\includegraphics[width=0.3\textwidth,height=0.2\textheight]{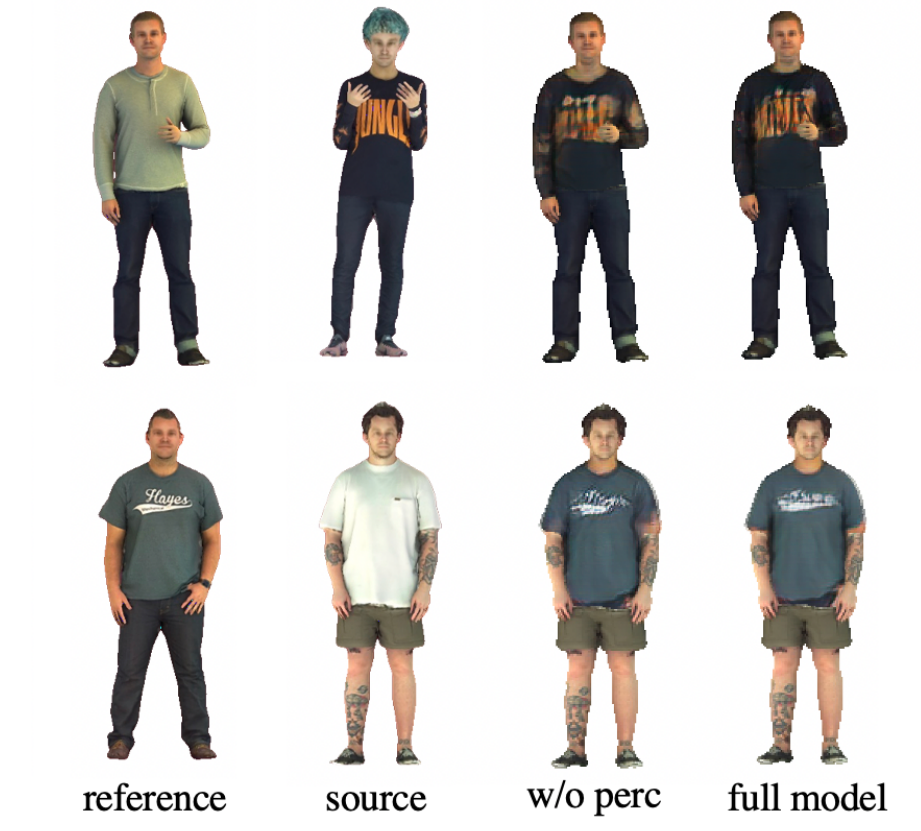}  
\caption{\textbf{Effects of the perceptual loss.} \textit{w/o perc} indicates that training our method without using perceptual loss.}
\label{11}
\end{figure}

\begin{figure}[htbp]
\centering 
\setlength{\abovecaptionskip}{0pt}
\includegraphics[width=0.48\textwidth,height=0.11\textheight]{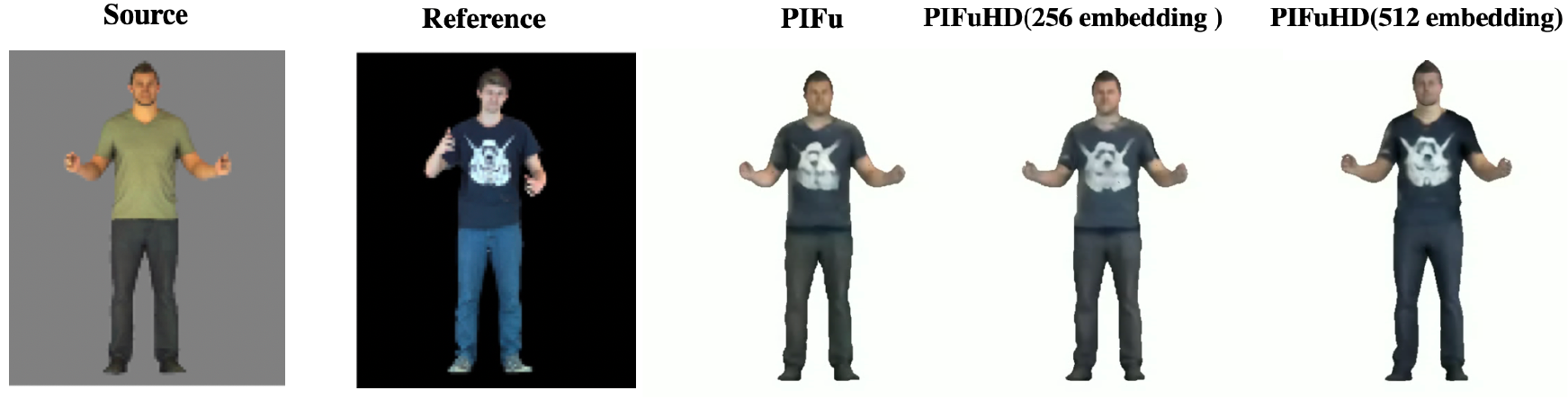}  
\caption{\textbf{Comparison of using PIFu and PIFuHD as baselines.} Note that we select different embeddings generated from PIFuHD\cite{saito2020pifuhd}.}
\label{12}
\end{figure}

\textbf{Effectiveness of the perceptual \& contextual \& cycle loss.} The pattern of the generated texture appears to be more sparse without the consideration of contextual loss in the fig. \ref{10}, which indicates that the use of contextual loss is important in generating high-quality textures with dense and intricate patterns that are visually appealing and realistic. The model trained with perceptual loss tends to transfer the accurate texture of both body and garments, which can be seen in Fig. \ref{11}. This means that using perceptual loss can improve the quality of the generated images, making them more realistic and visually appealing. As shown in Tab. \ref{2}, without using contextual loss or perceptual loss or cycle loss both makes SSIM scores higher while the FID and LPIPS scores lower. It's noticeable that the perceptual loss has the greatest impact on the results compared to other losses.

\textbf{Comparison between PIFu \& PIFuHD.} We leverage pixel-aligned features that combine spatial information proposed by PIFu\cite{saito2019pifu} to obtain both geometric shape and texture. PIFuhd \cite{saito2020pifuhd}, an improved version of PIFu, can generate higher resolution meshes. Since PIFuHD trains the reconstruction step using a coarse-to-fine approach without involving texture inference like PIFu, we adjusted the size of the input image during the coarse stage to obtain different sizes of geometric embeddings, which are 512 and 256, respectively, corresponding to input images of 1024 and 512. Then, we use geometric embeddings as the shape guidance to train texture. As shown in Fig. \ref{12}, the rendered texture using 512 embeddings through PIFuhd appears to be clearer. But to save on memory costs and simplify the processes, we chose to adapt PIFu instead of PIFuHD to generate the results demonstrated in this paper.

\vspace{1.0ex}
\noindent \textbf{Limitation:} When calculating the correspondence matrix in formula 2, it is often necessary to align two high-dimensional feature maps to find matching pixels. This process can be memory-intensive because it requires multiple copies of data to be stored in memory. To optimize memory consumption, we will explore alternative methods for calculating the similarity between two feature maps in future work.
\par
Most of the person poses in the MGN dataset are relatively simple, lacking samples with complex poses. In the 2D virtual try-on task, dressing for complex poses is a research hotspot. We also hope that our method will have robust performance on complex person poses, allowing us to expand our method to a wider range of applications. In the future, we plan to explore 3D human body datasets that focus on poses, and apply texture rendering to them for our task.
\par 
Although our method allows for flexible changes to any area of clothing, the texture we generate tends to be more averaged compared to the method of generating vertex-corresponding textures from pre-defined clothing templates. This is particularly noticeable in the back of the human body, which can appear blurry. This is a known drawback of PIFu's texture inference. NeRF\cite{mildenhall2021nerf} technology can synthesize clear and coherent new perspectives, and in the future we are considering combining our method with Nerf to obtain better textures.

\section{Conclusion}

Our proposed approach seamlessly reconstructs a 3D human mesh with the newly try-on result while preserving the texture from an arbitrary perspective. DI-Net comprises two modules: the complementary warping module, which leverages dense correspondence learning and sparse flow learning to warp the reference image to the same pose as the source image, and the geometry-aware decomposed transfer module, which decomposes the garment transfer into image layout-based transfer and texture-based transfer. This module achieves surface and texture reconstruction by constructing pixel-aligned implicit functions. Our experimental results demonstrate that DI-Net is highly effective and outperforms other existing methods in the 3D virtual try-on task. The generated textured mesh accurately captures the body appearance from the source image and preserves the consistent texture presented in the desired garments with geometry awareness. 

\section{Acknowledgement}
This work was supported by National Natural Science Foundation of China (NSFC) 62272172, Guangdong Basic and Applied Basic Research Foundation 2023A1515012920, Tip-top Scientific and Technical Innovative Youth Talents of Guangdong Special Support Program 2019TQ05X200 and 2022 Tencent Wechat Rhino-Bird Focused Research Program (Tencent WeChat RBFR2022008), and the Major Key Project of PCL under Grant PCL2021A09.


\bibliographystyle{IEEEtran}
\bibliography{IEEEabrv,mylib}

\begin{thebibliography}{10}
\providecommand{\url}[1]{#1}
\csname url@samestyle\endcsname
\providecommand{\newblock}{\relax}
\providecommand{\bibinfo}[2]{#2}
\providecommand{\BIBentrySTDinterwordspacing}{\spaceskip=0pt\relax}
\providecommand{\BIBentryALTinterwordstretchfactor}{4}
\providecommand{\BIBentryALTinterwordspacing}{\spaceskip=\fontdimen2\font plus
\BIBentryALTinterwordstretchfactor\fontdimen3\font minus
  \fontdimen4\font\relax}
\providecommand{\BIBforeignlanguage}[2]{{%
\expandafter\ifx\csname l@#1\endcsname\relax
\typeout{** WARNING: IEEEtran.bst: No hyphenation pattern has been}%
\typeout{** loaded for the language `#1'. Using the pattern for}%
\typeout{** the default language instead.}%
\else
\language=\csname l@#1\endcsname
\fi
#2}}
\providecommand{\BIBdecl}{\relax}
\BIBdecl

\bibitem{bhatnagar2020loopreg}
B.~L. Bhatnagar, C.~Sminchisescu, C.~Theobalt, and G.~Pons-Moll, ``Loopreg:
  Self-supervised learning of implicit surface correspondences, pose and shape
  for 3d human mesh registration,'' \emph{Advances in Neural Information
  Processing Systems}, vol.~33, pp. 12\,909--12\,922, 2020.

\bibitem{song20213d}
C.~Song, J.~Wei, R.~Li, F.~Liu, and G.~Lin, ``3d pose transfer with
  correspondence learning and mesh refinement,'' \emph{Advances in Neural
  Information Processing Systems}, vol.~34, 2021.

\bibitem{bhatnagar2020combining}
B.~L. Bhatnagar, C.~Sminchisescu, C.~Theobalt, and G.~Pons-Moll, ``Combining
  implicit function learning and parametric models for 3d human
  reconstruction,'' in \emph{European Conference on Computer Vision}.\hskip 1em
  plus 0.5em minus 0.4em\relax Springer, 2020, pp. 311--329.

\bibitem{tiwari2021neural}
G.~Tiwari, N.~Sarafianos, T.~Tung, and G.~Pons-Moll, ``Neural-gif: Neural
  generalized implicit functions for animating people in clothing,'' in
  \emph{Proceedings of the IEEE/CVF International Conference on Computer
  Vision}, 2021, pp. 11\,708--11\,718.

\bibitem{han2018viton}
X.~Han, Z.~Wu, Z.~Wu, R.~Yu, and L.~S. Davis, ``Viton: An image-based virtual
  try-on network,'' in \emph{Proceedings of the IEEE conference on computer
  vision and pattern recognition}, 2018, pp. 7543--7552.

\bibitem{dong2019towards}
H.~Dong, X.~Liang, X.~Shen, B.~Wang, H.~Lai, J.~Zhu, Z.~Hu, and J.~Yin,
  ``Towards multi-pose guided virtual try-on network,'' in \emph{Proceedings of
  the IEEE/CVF International Conference on Computer Vision}, 2019, pp.
  9026--9035.

\bibitem{yang2020towards}
H.~Yang, R.~Zhang, X.~Guo, W.~Liu, W.~Zuo, and P.~Luo, ``Towards
  photo-realistic virtual try-on by adaptively generating-preserving image
  content,'' in \emph{Proceedings of the IEEE/CVF Conference on Computer Vision
  and Pattern Recognition}, 2020, pp. 7850--7859.

\bibitem{yu2019vtnfp}
R.~Yu, X.~Wang, and X.~Xie, ``Vtnfp: An image-based virtual try-on network with
  body and clothing feature preservation,'' in \emph{Proceedings of the
  IEEE/CVF International Conference on Computer Vision}, 2019, pp.
  10\,511--10\,520.

\bibitem{wang2018toward}
B.~Wang, H.~Zheng, X.~Liang, Y.~Chen, L.~Lin, and M.~Yang, ``Toward
  characteristic-preserving image-based virtual try-on network,'' in
  \emph{Proceedings of the European Conference on Computer Vision (ECCV)},
  2018, pp. 589--604.

\bibitem{guan2012drape}
P.~Guan, L.~Reiss, D.~A. Hirshberg, A.~Weiss, and M.~J. Black, ``Drape:
  Dressing any person,'' \emph{ACM Transactions on Graphics (TOG)}, vol.~31,
  no.~4, pp. 1--10, 2012.

\bibitem{hahn2014subspace}
F.~Hahn, B.~Thomaszewski, S.~Coros, R.~W. Sumner, F.~Cole, M.~Meyer, T.~DeRose,
  and M.~Gross, ``Subspace clothing simulation using adaptive bases,''
  \emph{ACM Transactions on Graphics (TOG)}, vol.~33, no.~4, pp. 1--9, 2014.

\bibitem{lahner2018deepwrinkles}
Z.~Lahner, D.~Cremers, and T.~Tung, ``Deepwrinkles: Accurate and realistic
  clothing modeling,'' in \emph{Proceedings of the European Conference on
  Computer Vision (ECCV)}, 2018, pp. 667--684.

\bibitem{pons2017clothcap}
G.~Pons-Moll, S.~Pujades, S.~Hu, and M.~J. Black, ``Clothcap: Seamless 4d
  clothing capture and retargeting,'' \emph{ACM Transactions on Graphics
  (ToG)}, vol.~36, no.~4, pp. 1--15, 2017.

\bibitem{bhatnagar2019multi}
B.~L. Bhatnagar, G.~Tiwari, C.~Theobalt, and G.~Pons-Moll, ``Multi-garment net:
  Learning to dress 3d people from images,'' in \emph{Proceedings of the
  IEEE/CVF international conference on computer vision}, 2019, pp. 5420--5430.

\bibitem{ma2020learning}
Q.~Ma, J.~Yang, A.~Ranjan, S.~Pujades, G.~Pons-Moll, S.~Tang, and M.~J. Black,
  ``Learning to dress 3d people in generative clothing,'' in \emph{Proceedings
  of the IEEE/CVF Conference on Computer Vision and Pattern Recognition}, 2020,
  pp. 6469--6478.

\bibitem{mir2020learning}
A.~Mir, T.~Alldieck, and G.~Pons-Moll, ``Learning to transfer texture from
  clothing images to 3d humans,'' in \emph{Proceedings of the IEEE/CVF
  Conference on Computer Vision and Pattern Recognition}, 2020, pp. 7023--7034.

\bibitem{loper2015smpl}
M.~Loper, N.~Mahmood, J.~Romero, G.~Pons-Moll, and M.~J. Black, ``Smpl: A
  skinned multi-person linear model,'' \emph{ACM transactions on graphics
  (TOG)}, vol.~34, no.~6, pp. 1--16, 2015.

\bibitem{zhao2021m3d}
F.~Zhao, Z.~Xie, M.~Kampffmeyer, H.~Dong, S.~Han, T.~Zheng, T.~Zhang, and
  X.~Liang, ``M3d-vton: A monocular-to-3d virtual try-on network,'' in
  \emph{Proceedings of the IEEE/CVF International Conference on Computer
  Vision}, 2021, pp. 13\,239--13\,249.

\bibitem{zhu2019detailed}
H.~Zhu, X.~Zuo, S.~Wang, X.~Cao, and R.~Yang, ``Detailed human shape estimation
  from a single image by hierarchical mesh deformation,'' in \emph{Proceedings
  of the IEEE/CVF Conference on Computer Vision and Pattern Recognition}, 2019,
  pp. 4491--4500.

\bibitem{mildenhall2021nerf}
B.~Mildenhall, P.~P. Srinivasan, M.~Tancik, J.~T. Barron, R.~Ramamoorthi, and
  R.~Ng, ``Nerf: Representing scenes as neural radiance fields for view
  synthesis,'' \emph{Communications of the ACM}, vol.~65, no.~1, pp. 99--106,
  2021.

\bibitem{zheng2021pamir}
Z.~Zheng, T.~Yu, Y.~Liu, and Q.~Dai, ``Pamir: Parametric model-conditioned
  implicit representation for image-based human reconstruction,'' \emph{IEEE
  transactions on pattern analysis and machine intelligence}, vol.~44, no.~6,
  pp. 3170--3184, 2021.

\bibitem{chibane2020implicit}
J.~Chibane, T.~Alldieck, and G.~Pons-Moll, ``Implicit functions in feature
  space for 3d shape reconstruction and completion,'' in \emph{Proceedings of
  the IEEE/CVF Conference on Computer Vision and Pattern Recognition}, 2020,
  pp. 6970--6981.

\bibitem{mescheder2019occupancy}
L.~Mescheder, M.~Oechsle, M.~Niemeyer, S.~Nowozin, and A.~Geiger, ``Occupancy
  networks: Learning 3d reconstruction in function space,'' in
  \emph{Proceedings of the IEEE/CVF Conference on Computer Vision and Pattern
  Recognition}, 2019, pp. 4460--4470.

\bibitem{park2019deepsdf}
J.~J. Park, P.~Florence, J.~Straub, R.~Newcombe, and S.~Lovegrove, ``Deepsdf:
  Learning continuous signed distance functions for shape representation,'' in
  \emph{Proceedings of the IEEE/CVF Conference on Computer Vision and Pattern
  Recognition}, 2019, pp. 165--174.

\bibitem{du2022vton}
C.~Du, F.~Yu, M.~Jiang, A.~Hua, X.~Wei, T.~Peng, and X.~Hu, ``Vton-scfa: A
  virtual try-on network based on the semantic constraints and flow
  alignment,'' \emph{IEEE Transactions on Multimedia}, 2022.

\bibitem{hu2022spg}
B.~Hu, P.~Liu, Z.~Zheng, and M.~Ren, ``Spg-vton: Semantic prediction guidance
  for multi-pose virtual try-on,'' \emph{IEEE Transactions on Multimedia},
  vol.~24, pp. 1233--1246, 2022.

\bibitem{xu2021virtual}
J.~Xu, Y.~Pu, R.~Nie, D.~Xu, Z.~Zhao, and W.~Qian, ``Virtual try-on network
  with attribute transformation and local rendering,'' \emph{IEEE Transactions
  on Multimedia}, vol.~23, pp. 2222--2234, 2021.

\bibitem{bookstein1991thin}
F.~L. Bookstein, ``Thin-plate splines and the atlas problem for biomedical
  images,'' in \emph{Biennial international conference on information
  processing in medical imaging}.\hskip 1em plus 0.5em minus 0.4em\relax
  Springer, 1991, pp. 326--342.

\bibitem{raj2018swapnet}
A.~Raj, P.~Sangkloy, H.~Chang, J.~Hays, D.~Ceylan, and J.~Lu, ``Swapnet: Image
  based garment transfer,'' in \emph{European Conference on Computer
  Vision}.\hskip 1em plus 0.5em minus 0.4em\relax Springer, 2018, pp. 679--695.

\bibitem{neuberger2020image}
A.~Neuberger, E.~Borenstein, B.~Hilleli, E.~Oks, and S.~Alpert, ``Image based
  virtual try-on network from unpaired data,'' in \emph{Proceedings of the
  IEEE/CVF Conference on Computer Vision and Pattern Recognition}, 2020, pp.
  5184--5193.

\bibitem{wu2019m2e}
Z.~Wu, G.~Lin, Q.~Tao, and J.~Cai, ``M2e-try on net: Fashion from model to
  everyone,'' in \emph{Proceedings of the 27th ACM International Conference on
  Multimedia}, 2019, pp. 293--301.

\bibitem{zhong2021mv}
X.~Zhong, Z.~Wu, T.~Tan, G.~Lin, and Q.~Wu, ``Mv-ton: Memory-based video
  virtual try-on network,'' in \emph{Proceedings of the 29th ACM International
  Conference on Multimedia}, 2021, pp. 908--916.

\bibitem{men2020controllable}
Y.~Men, Y.~Mao, Y.~Jiang, W.-Y. Ma, and Z.~Lian, ``Controllable person image
  synthesis with attribute-decomposed gan,'' in \emph{Proceedings of the
  IEEE/CVF Conference on Computer Vision and Pattern Recognition}, 2020, pp.
  5084--5093.

\bibitem{yang2021ct}
F.~Yang and G.~Lin, ``Ct-net: Complementary transfering network for garment
  transfer with arbitrary geometric changes,'' in \emph{Proceedings of the
  IEEE/CVF Conference on Computer Vision and Pattern Recognition}, 2021, pp.
  9899--9908.

\bibitem{cui2021dressing}
A.~Cui, D.~McKee, and S.~Lazebnik, ``Dressing in order: Recurrent person image
  generation for pose transfer, virtual try-on and outfit editing,'' in
  \emph{Proceedings of the IEEE/CVF International Conference on Computer
  Vision}, 2021, pp. 14\,638--14\,647.

\bibitem{jiang2020bcnet}
B.~Jiang, J.~Zhang, Y.~Hong, J.~Luo, L.~Liu, and H.~Bao, ``Bcnet: Learning body
  and cloth shape from a single image,'' in \emph{European Conference on
  Computer Vision}.\hskip 1em plus 0.5em minus 0.4em\relax Springer, 2020, pp.
  18--35.

\bibitem{patel2020tailornet}
C.~Patel, Z.~Liao, and G.~Pons-Moll, ``Tailornet: Predicting clothing in 3d as
  a function of human pose, shape and garment style,'' in \emph{Proceedings of
  the IEEE/CVF Conference on Computer Vision and Pattern Recognition}, 2020,
  pp. 7365--7375.

\bibitem{santesteban2021self}
I.~Santesteban, N.~Thuerey, M.~A. Otaduy, and D.~Casas, ``Self-supervised
  collision handling via generative 3d garment models for virtual try-on,'' in
  \emph{Proceedings of the IEEE/CVF Conference on Computer Vision and Pattern
  Recognition}, 2021, pp. 11\,763--11\,773.

\bibitem{wang2020normalgan}
L.~Wang, X.~Zhao, T.~Yu, S.~Wang, and Y.~Liu, ``Normalgan: Learning detailed 3d
  human from a single rgb-d image,'' in \emph{European Conference on Computer
  Vision}.\hskip 1em plus 0.5em minus 0.4em\relax Springer, 2020, pp. 430--446.

\bibitem{danvevrek2017deepgarment}
R.~Dan{\v{e}}{\v{r}}ek, E.~Dibra, C.~{\"O}ztireli, R.~Ziegler, and M.~Gross,
  ``Deepgarment: 3d garment shape estimation from a single image,'' in
  \emph{Computer Graphics Forum}, vol.~36, no.~2.\hskip 1em plus 0.5em minus
  0.4em\relax Wiley Online Library, 2017, pp. 269--280.

\bibitem{gundogdu2019garnet}
E.~Gundogdu, V.~Constantin, A.~Seifoddini, M.~Dang, M.~Salzmann, and P.~Fua,
  ``Garnet: A two-stream network for fast and accurate 3d cloth draping,'' in
  \emph{Proceedings of the IEEE/CVF International Conference on Computer
  Vision}, 2019, pp. 8739--8748.

\bibitem{santesteban2019learning}
I.~Santesteban, M.~A. Otaduy, and D.~Casas, ``Learning-based animation of
  clothing for virtual try-on,'' in \emph{Computer Graphics Forum}, vol.~38,
  no.~2.\hskip 1em plus 0.5em minus 0.4em\relax Wiley Online Library, 2019, pp.
  355--366.

\bibitem{xiang2020monoclothcap}
D.~Xiang, F.~Prada, C.~Wu, and J.~Hodgins, ``Monoclothcap: Towards temporally
  coherent clothing capture from monocular rgb video,'' in \emph{2020
  International Conference on 3D Vision (3DV)}.\hskip 1em plus 0.5em minus
  0.4em\relax IEEE, 2020, pp. 322--332.

\bibitem{tiwari2020sizer}
G.~Tiwari, B.~L. Bhatnagar, T.~Tung, and G.~Pons-Moll, ``Sizer: A dataset and
  model for parsing 3d clothing and learning size sensitive 3d clothing,'' in
  \emph{European Conference on Computer Vision}.\hskip 1em plus 0.5em minus
  0.4em\relax Springer, 2020, pp. 1--18.

\bibitem{pavlakos2019expressive}
G.~Pavlakos, V.~Choutas, N.~Ghorbani, T.~Bolkart, A.~A. Osman, D.~Tzionas, and
  M.~J. Black, ``Expressive body capture: 3d hands, face, and body from a
  single image,'' in \emph{Proceedings of the IEEE/CVF conference on computer
  vision and pattern recognition}, 2019, pp. 10\,975--10\,985.

\bibitem{ma2021power}
Q.~Ma, J.~Yang, S.~Tang, and M.~J. Black, ``The power of points for modeling
  humans in clothing,'' in \emph{Proceedings of the IEEE/CVF International
  Conference on Computer Vision}, 2021, pp. 10\,974--10\,984.

\bibitem{ma2021scale}
Q.~Ma, S.~Saito, J.~Yang, S.~Tang, and M.~J. Black, ``Scale: Modeling clothed
  humans with a surface codec of articulated local elements,'' in
  \emph{Proceedings of the IEEE/CVF Conference on Computer Vision and Pattern
  Recognition}, 2021, pp. 16\,082--16\,093.

\bibitem{saito2019pifu}
S.~Saito, Z.~Huang, R.~Natsume, S.~Morishima, A.~Kanazawa, and H.~Li, ``Pifu:
  Pixel-aligned implicit function for high-resolution clothed human
  digitization,'' in \emph{Proceedings of the IEEE/CVF International Conference
  on Computer Vision}, 2019, pp. 2304--2314.

\bibitem{saito2020pifuhd}
S.~Saito, T.~Simon, J.~Saragih, and H.~Joo, ``Pifuhd: Multi-level pixel-aligned
  implicit function for high-resolution 3d human digitization,'' in
  \emph{Proceedings of the IEEE/CVF Conference on Computer Vision and Pattern
  Recognition}, 2020, pp. 84--93.

\bibitem{cao2017realtime}
Z.~Cao, T.~Simon, S.-E. Wei, and Y.~Sheikh, ``Realtime multi-person 2d pose
  estimation using part affinity fields,'' in \emph{Proceedings of the IEEE
  conference on computer vision and pattern recognition}, 2017, pp. 7291--7299.

\bibitem{zhang2020cross}
P.~Zhang, B.~Zhang, D.~Chen, L.~Yuan, and F.~Wen, ``Cross-domain correspondence
  learning for exemplar-based image translation,'' in \emph{Proceedings of the
  IEEE/CVF Conference on Computer Vision and Pattern Recognition}, 2020, pp.
  5143--5153.

\bibitem{zhang2019deep}
B.~Zhang, M.~He, J.~Liao, P.~V. Sander, L.~Yuan, A.~Bermak, and D.~Chen, ``Deep
  exemplar-based video colorization,'' in \emph{Proceedings of the IEEE/CVF
  Conference on Computer Vision and Pattern Recognition}, 2019, pp. 8052--8061.

\bibitem{luo2018cosine}
C.~Luo, J.~Zhan, X.~Xue, L.~Wang, R.~Ren, and Q.~Yang, ``Cosine normalization:
  Using cosine similarity instead of dot product in neural networks,'' in
  \emph{International Conference on Artificial Neural Networks}.\hskip 1em plus
  0.5em minus 0.4em\relax Springer, 2018, pp. 382--391.

\bibitem{wei2017learning}
Z.~Wei, Y.~Sun, J.~Wang, H.~Lai, and S.~Liu, ``Learning adaptive receptive
  fields for deep image parsing network,'' in \emph{Proceedings of the IEEE
  conference on computer vision and pattern recognition}, 2017, pp. 2434--2442.

\bibitem{ren2019structureflow}
Y.~Ren, X.~Yu, R.~Zhang, T.~H. Li, S.~Liu, and G.~Li, ``Structureflow: Image
  inpainting via structure-aware appearance flow,'' in \emph{Proceedings of the
  IEEE/CVF International Conference on Computer Vision}, 2019, pp. 181--190.

\bibitem{park2019semantic}
T.~Park, M.-Y. Liu, T.-C. Wang, and J.-Y. Zhu, ``Semantic image synthesis with
  spatially-adaptive normalization,'' in \emph{Proceedings of the IEEE/CVF
  conference on computer vision and pattern recognition}, 2019, pp. 2337--2346.

\bibitem{wang2018high}
T.-C. Wang, M.-Y. Liu, J.-Y. Zhu, A.~Tao, J.~Kautz, and B.~Catanzaro,
  ``High-resolution image synthesis and semantic manipulation with conditional
  gans,'' in \emph{Proceedings of the IEEE conference on computer vision and
  pattern recognition}, 2018, pp. 8798--8807.

\bibitem{lorensen1987marching}
W.~E. Lorensen and H.~E. Cline, ``Marching cubes: A high resolution 3d surface
  construction algorithm,'' \emph{ACM siggraph computer graphics}, vol.~21,
  no.~4, pp. 163--169, 1987.

\bibitem{johnson2016perceptual}
J.~Johnson, A.~Alahi, and L.~Fei-Fei, ``Perceptual losses for real-time style
  transfer and super-resolution,'' in \emph{European conference on computer
  vision}.\hskip 1em plus 0.5em minus 0.4em\relax Springer, 2016, pp. 694--711.

\bibitem{simonyan2014very}
K.~Simonyan and A.~Zisserman, ``Very deep convolutional networks for
  large-scale image recognition,'' \emph{arXiv preprint arXiv:1409.1556}, 2014.

\bibitem{mechrez2018contextual}
R.~Mechrez, I.~Talmi, and L.~Zelnik-Manor, ``The contextual loss for image
  transformation with non-aligned data,'' in \emph{Proceedings of the European
  conference on computer vision (ECCV)}, 2018, pp. 768--783.

\bibitem{mirza2014conditional}
M.~Mirza and S.~Osindero, ``Conditional generative adversarial nets,''
  \emph{arXiv preprint arXiv:1411.1784}, 2014.

\bibitem{renderpeople}
\BIBentryALTinterwordspacing
``Renderpeople. 2018.'' [Online]. Available: \url{https://renderpeople.com}
\BIBentrySTDinterwordspacing

\bibitem{zheng2019deephuman}
Z.~Zheng, T.~Yu, Y.~Wei, Q.~Dai, and Y.~Liu, ``Deephuman: 3d human
  reconstruction from a single image,'' in \emph{Proceedings of the IEEE/CVF
  International Conference on Computer Vision}, 2019, pp. 7739--7749.

\bibitem{yu2021function4d}
T.~Yu, Z.~Zheng, K.~Guo, P.~Liu, Q.~Dai, and Y.~Liu, ``Function4d: Real-time
  human volumetric capture from very sparse consumer rgbd sensors,'' in
  \emph{Proceedings of the IEEE/CVF Conference on Computer Vision and Pattern
  Recognition}, 2021, pp. 5746--5756.

\bibitem{varol2017learning}
G.~Varol, J.~Romero, X.~Martin, N.~Mahmood, M.~J. Black, I.~Laptev, and
  C.~Schmid, ``Learning from synthetic humans,'' in \emph{Proceedings of the
  IEEE conference on computer vision and pattern recognition}, 2017, pp.
  109--117.

\bibitem{liu2016deepfashion}
Z.~Liu, P.~Luo, S.~Qiu, X.~Wang, and X.~Tang, ``Deepfashion: Powering robust
  clothes recognition and retrieval with rich annotations,'' in
  \emph{Proceedings of the IEEE conference on computer vision and pattern
  recognition}, 2016, pp. 1096--1104.

\bibitem{jaderberg2015spatial}
M.~Jaderberg, K.~Simonyan, A.~Zisserman \emph{et~al.}, ``Spatial transformer
  networks,'' \emph{Advances in neural information processing systems},
  vol.~28, 2015.

\bibitem{wang2004image}
Z.~Wang, A.~C. Bovik, H.~R. Sheikh, and E.~P. Simoncelli, ``Image quality
  assessment: from error visibility to structural similarity,'' \emph{IEEE
  transactions on image processing}, vol.~13, no.~4, pp. 600--612, 2004.

\bibitem{heusel2017gans}
M.~Heusel, H.~Ramsauer, T.~Unterthiner, B.~Nessler, and S.~Hochreiter, ``Gans
  trained by a two time-scale update rule converge to a local nash
  equilibrium,'' \emph{Advances in neural information processing systems},
  vol.~30, 2017.

\bibitem{zhang2018unreasonable}
R.~Zhang, P.~Isola, A.~A. Efros, E.~Shechtman, and O.~Wang, ``The unreasonable
  effectiveness of deep features as a perceptual metric,'' in \emph{Proceedings
  of the IEEE conference on computer vision and pattern recognition}, 2018, pp.
  586--595.

\bibitem{zhou2021semantic}
H.~Zhou, W.~Wu, Y.~Zhang, J.~Ma, and H.~Ling, ``Semantic-supervised infrared
  and visible image fusion via a dual-discriminator generative adversarial
  network,'' \emph{IEEE Transactions on Multimedia}, 2021.

\bibitem{pang2021image}
Y.~Pang, J.~Lin, T.~Qin, and Z.~Chen, ``Image-to-image translation: Methods and
  applications,'' \emph{IEEE Transactions on Multimedia}, vol.~24, pp.
  3859--3881, 2021.

\bibitem{lin2023smnet}
S.~Lin, F.~Tang, W.~Dong, X.~Pan, and C.~Xu, ``Smnet: Synchronous multi-scale
  low light enhancement network with local and global concern,'' \emph{IEEE
  Transactions on Multimedia}, 2023.

\end{thebibliography}


\end{document}